\documentclass[sagev,Afour,times]{sagej}

\usepackage{moreverb,url}

\usepackage[colorlinks,bookmarksopen,bookmarksnumbered,citecolor=red,urlcolor=red]{hyperref}

\newcommand\BibTeX{{\rmfamily B\kern-.05em \textsc{i\kern-.025em b}\kern-.08em
T\kern-.1667em\lower.7ex\hbox{E}\kern-.125emX}}

\def\volumeyear{2021}

\usepackage{graphicx}                

\usepackage{textcomp}                  
\usepackage{mathptmx}                  
\usepackage{amsfonts}
\usepackage{amsmath}
\usepackage{rotating}
\usepackage[T1]{fontenc}
\usepackage{tabularx,booktabs}
\usepackage{multirow}
\usepackage{xcolor}
\usepackage{microtype}

\begin{document}

\runninghead{Kim \emph{et al.}}

\title{Visual Cluster Separation Using High-Dimensional Sharpened Dimensionality Reduction}

\author{Youngjoo~Kim\affilnum{1},
        Alexandru~C.~Telea\affilnum{2},
         Scott~C.~Trager\affilnum{3},
         and~Jos~B.~T.~M.~Roerdink\affilnum{1}}

\affiliation{\affilnum{1}Bernoulli Institute for Mathematics, Computer Science and Artificial Intelligence, University of Groningen, The Netherlands.\\
\affilnum{2}Department of Information and Computing Sciences, Utrecht University, The Netherlands.\\
\affilnum{3}Kapteyn Astronomical Institute, University of Groningen, The Netherlands.}

\corrauth{Youngjoo Kim,
Bernoulli Institute,
University of Groningen,
Nijenborgh 9,
9747 AG, Netherlands.}
\email{lyoungjookiml@gmail.com}

\begin{abstract}
Applying dimensionality reduction (DR) to large, high-dimensional data sets can be challenging when distinguishing the underlying high-dimensional data clusters in a 2D projection for exploratory analysis. We address this problem by first sharpening the clusters in the original high-dimensional data prior to the DR step using Local Gradient Clustering (LGC). We then project the sharpened data from the high-dimensional space to 2D by a user-selected DR method. The sharpening step aids this method to preserve cluster separation in the resulting 2D projection. With our method, end-users can label each distinct cluster to further analyze an otherwise unlabeled data set. Our `High-Dimensional Sharpened DR' (HD-SDR) method, tested on both synthetic and real-world data sets, is favorable to DR methods with poor cluster separation and yields a better visual cluster separation than these DR methods with no sharpening. Our method achieves good quality (measured by quality metrics) and scales computationally well with large high-dimensional data. To illustrate its concrete applications, we further apply HD-SDR on a recent astronomical catalog.
\end{abstract}

\keywords{High-dimensional data visualization, dimensionality reduction, clustering, astronomy}

\maketitle

\section{Introduction}
\label{sec:intro}
Dimensionality reduction (DR) techniques depict high-dimensional data with low-dimensional scatter plots. DR is widely used because it preserves the structure of high-dimensional data. For example, when the data is distributed over several clusters, DR allows one to directly and visually examine such structures in 2D or 3D, in terms of visually well-separated point clusters in a scatterplot. While \emph{t}-distributed Stochastic Neighbor Embedding (\emph{t}-SNE\,\cite{tsne:original}) is arguably one of the best DR techniques in creating visually well-separated clusters of similar-data points, the recent work of Anders \emph{et al.}\ shows that even with \emph{t}-SNE, when visual clusters overlap even slightly, manually labeling them can be challenging\,\cite{tsne:astro}. Besides \emph{t}-SNE, many other nonlinear DR techniques have been proposed, \emph{e.g.}, Random Projection (RP)\,\cite{RP1:randomProj,rp2:randomproj_survey}, Landmark Multidimensional Scaling (LMDS)\,\cite{lmds:original}, ISOMAP\,\cite{isomap:original}, Sammon Mapping\,\cite{sammonMapping:original}, and Uniform Manifold Approximation and Projection (UMAP)\,\cite{umap2018}. While such methods typically achieve a poorer visual cluster separation than \emph{t}-SNE\,\cite{dr:review,tsne:original}, they are computationally more scalable and simpler to implement and use\,\cite{mateusDR_survey2019}.

Espadoto \emph{et al.}\,\cite{mateusDR_survey2019} have benchmarked dozens of DR techniques using several quality metrics and showed that there is no `ideal' DR technique that guarantees the visual separation of similar-data clusters for any kind of data. As such, we are interested to find a generic approach to \emph{improve} upon existing DR methods in terms of visual cluster separation while keeping other attractive specific features these already have, \emph{e.g.}, neighborhood and distance preservation, computational scalability, or simplicity.

In this paper, we show how sharpening the clusters in the original high-dimensional data can enhance Visual Cluster Separation (VCS) -- loosely defined, for now, as the ability of a user to see separate clusters in a 2D projection. A more formal definition and explanation of the importance of VCS is introduced in Section~\nameref{sec:relatedwork:VCS}. We sharpen the data clusters by Local Gradient Clustering (LGC) and then project the sharpened data to 2D using standard DR techniques. When the input high-dimensional data has cluster structures, our
`High-Dimensional Sharpened DR' (HD-SDR) method creates projections that show these clusters more clearly and better separated from each other than when using the baseline, original, DR method alone. As such, our approach is not a new DR technique, but a new way to enhance the VCS properties of any existing DR technique.
To our knowledge, this the first time that such a sharpening approach is used to enhance VCS without any prior estimation of cluster modes\,\cite{realworld:banknote_sep}. 

We evaluate our HD-SDR method on synthetic and real-world labeled data using quality metrics that empirically and theoretically measure the preservation of neighbors and their corresponding labels, and use RP, LMDS, \emph{t}-SNE, and UMAP as the baseline DR methods\,\cite{RP1:randomProj,lmds:original,tsne:original,umap2018}. By comparing the baseline DR with HD-SDR, our results show that sharpening assists those DR methods, which have difficulty in producing visually well-separated clusters, and create projections with clear VCS.

To demonstrate the practical usefulness of HD-SDR, we apply it to explore an unlabeled real-world astronomical data set drawn from the recent GALactic Archaeology with HERMES Data Release 2 (GALAH DR2) and \emph{Gaia} Data Release 2 (Gaia DR2) catalogues\,\cite{astro:GAIADR2_1, astro:GAIADR2_2, astro:GALAHDR2, tsne:astro}. Astronomers are able to label and further analyze each distinct cluster using our method. This use-case shows how our method can easily assist domain experts to manually and visually label data clusters by annotating their 2D projections, which leads to a better understanding of the large high-dimensional data at hand. Although currently out of our scope, HD-SDR could further be used to assist user-guided labeling in semi-supervised learning, where small portions of labeled data (given or manually assigned by end-users) are used to propagate labels to unlabeled data, which are then used to train a conventional classifier\,\cite{label_propagation_wang2007,semisupervised_clustering_userfeedback2003,barbara2018,benato21,labeling_wt_DR2017}.

In summary, our contributions are as follows:

\begin{itemize}
    \item We propose a novel method to improve Visual Cluster Separation (VCS) of DR methods by sharpening the original high-dimensional data prior to the projection. This is to our knowledge the first time that such a sharpening method is used to improve VCS in DR without any prior estimation of the cluster modes;
    \item We demonstrate both qualitatively and quantitatively that our method enhances VCS for DR methods that originally show weak cluster separation;
    \item We apply our method to unlabeled real-world astronomical data and show evidence that the resulting visual clusters have a physical meaning in our Milky Way Galaxy.
\end{itemize}

This paper is structured as follows. Section~\nameref{sec:relatedWork} outlines related work in dimensionality reduction. Section~\nameref{sec:method} details our method. Section~\nameref{sec:results} compares our method qualitatively and quantitatively with standard DR on several synthetic and real-world data sets. Section~\nameref{sec:astro} shows a practical use-case with unlabeled real-world astronomical data. Section~\nameref{sec:discussion} discusses several aspects of our method, including its cluster segregation power, data distortion, scalability, and limitations. Section~\nameref{sec:conclusion} concludes the paper.

\section{Related Work} \label{sec:relatedWork}
We first briefly discuss the relation between cluster separation and DR used for exploratory analysis in Section~\nameref{sec:relatedwork:VCS}. We next explain the importance of cluster separation in DR used for data labeling (Section~\nameref{sec:relatedwork:DRforLabeling}), followed by specific use-cases of DR in astronomy, our main application area (Section~\nameref{sec:relatedwork:astro}). 
    
\subsection{Dimensionality Reduction and Cluster Separation}
\label{sec:relatedwork:VCS}
While DR has multiple goals such as data compression\,\cite{ae_survey}, feature extraction\,\cite{bleha91}, and exploratory analysis\,\cite{nonato18}, we focus here on the exploratory analysis using DR methods to visually support identifying clusters of similar-value data points. Finding separate clusters, defined as sets of unlabeled data points that have similarities but are different from other point sets, is challenging in data science and unsupervised learning. Data clustering serves multiple aims: \emph{e.g.}, finding natural modes or types of samples in distributions; classification; data aggregation and simplification; and data visualization. Although clustering algorithms do not explicitly use predefined labels as in supervised learning, they still need \emph{a priori} knowledge of the data. For instance, $k$-means clustering explicitly requires the number of clusters\,\cite{kmeans1967}; hierarchical clustering requires defining a similarity threshold\,\cite{hierarchicalClust}; and DBSCAN asks for the minimum number of neighborhoods required to form a dense region\,\cite{dbscan1972, cluster_berkhin2006survey}. Given the above, there is no unique and/or `correct' clustering of a given data set. Instead, the `cluster structure' present in a data set is implied by a given clustering method and its hyperparameters.

Let $D = \{\mathbf{x}_1,\ldots,\mathbf{x}_N\}$ be a set of $n$-dimensional observations (samples, points), $\mathbf{x}_i = [x^1_i$ $x^2_i$ $\cdots$ $x^n_i] \in \mathbb{R}^n$. Here, $x^j_i$ ($1 \leq j \leq n$) is the $j^{\text{th}}$ attribute value of the $i^{th}$ sample. A DR technique, or projection, can be modeled by a function $P: \mathbb{R}^n \rightarrow \mathbb{R}^s$. In practice $s=2$ is most used -- for details we refer to Martins\,\cite{dr:definition}. 
A projection function $P$ allows one to reason about a data set $D \subset \mathbb{R}^n$ by visually interpreting its projection (scatterplot), which we denote as $P(D) = \{ P(\mathbf{x}_i) | \mathbf{x}_i \in D\}$. Hence, if data structure in terms of clusters exists in $D$ (in the sense outlined in the previous paragraph), these should also be visible in $P(D)$. A projection $P$ reflects the \emph{data cluster separation} present in $D$ by the \emph{visual cluster separation} present in $P(D)$\,\cite{rauber,rauber_2}. Note that the function $P$ from $\mathbb{R}^n$ to $\mathbb{R}^2$ is, in general, many-to-one in terms of point locations, \emph{i.e.}, points that have different coordinates in $\mathbb{R}^n$ can be mapped to the same location in $\mathbb{R}^2$. Yet, every sample point $\mathbf{x}_i \in D$ is mapped to a unique point $P(\mathbf{x}_i) \in P(D)$ by using the index $i$ as an identifier both in $D$ and $P(D)$ For these reasons, we use the term `labeling' to refer to adding class labels to either the 2D or the $n$D data sets, as labeling a point $P(\mathbf{x}_i) \in P(D)$ in the projection directly labels its corresponding point $\mathbf{x}_i \in D$ in the data set, and conversely.

We can define (visual) cluster separation more formally as follows. Let $H : S \rightarrow X$ be any metric or tool that is able to reason about the clusters present in a data set $S$. Let $H(S)$ denote the application of $H$ to all points in $S$. When $S = D$, $H$ captures the \emph{data cluster separation}. When $S = P(D)$, $H$ captures \emph{visual cluster separation (VCS)}. Examples of $H(D)$ are classifiers that assign a label $x \in X$ to data points so that points in the same cluster get the same label; clustering techniques that assign a cluster ID to similar data points (in this case $X \subset \mathbb{N}$) or count the number of clusters in a data set (in this case, $X = \mathbb{N}$). Several instances of $H(P(D))$ have been proposed to measure VCS in visualization research\,\cite{mateusDR_survey2019}. Given the above, we say that a projection $P$ has good VCS when $H(P(D))$ is very similar to $H(D)$, \emph{i.e.}, $P$ should ideally capture in the 2D visual space the same cluster structure that the metric $H$ finds in the data space. Note that `good VCS' does not identically mean `high VCS'. Rather, good VCS implies two cases: (a) When $H(D)$ is high (data is well separated in the high-dimensional space), then $H(P(D))$ should also be high; and (b) when $H(D)$ is low (there is no clear cluster structure in the data), then $H(P(D))$ should also be low (the projection should not create artificial visual clusters that wrongly suggest that the data has this type of structure). Following this, there are two cases when $H(P(D))$ does not reflect well $H(D)$: we say that (1) $P$ \emph{undersegments} the data $D$ if $H(D)$ contains more clusters than $H(P(D))$; this can be seen as $P(D)$ showing `false negatives' in terms of missing visual clusters; and (2) $P$ \emph{oversegments} $D$ if $H(D)$ contains fewer clusters than $H(P(D))$; this can be seen as $P(D)$ showing `false positives' in terms of spurious visual clusters.

Visual cluster separation in distance-preserving projections of intrinsically low-dimensional data, where the $n$D distances are reflected well by their corresponding 2D counterparts, is easier to spot based on the distances between clusters.
However, we argue that these cases of intrinsically low-dimensional data embedded into high-dimensional spaces are a minority. When exploring high-dimensional data by non-linear projections or projections that do not preserve distances (but neighborhoods), looking at visual clusters in $P(D)$ is the only way to reason about $D$ because the exact inter-point distances in $P(D)$ have little meaning. Hence, for such projections, VCS is also important. This is also reflected in methods such as t-SNE and UMAP\,\cite{tsne:original,umap2018}.

\subsection{Dimensionality reduction for labeling} \label{sec:relatedwork:DRforLabeling}
Semi-supervised learning methods propagate labels from a small set of predefined labeled data to the remaining unlabeled data points prior to training a conventional classifier\,\cite{label_propagation_wang2007, semisupervised_clustering_userfeedback2003, barbara2018,benato21}. These methods take advantage of DR by letting users assign or propagate labels directly, and visually, in a projection. In visual analytics, user-centered Visual-Interactive Labeling (VIL) is combined with model-centered Active Learning (AL) to achieve better labeling\,\cite{labeling_wt_DR2017}. Such methods are highly effective when not enough labeled training data exist and/or when users need more control on label propagation. Yet, VIL requires strong VCS so users know when to stop visually propagating a label\,\cite{barbara2018,benato21}, which not all DR methods deliver, as mentioned earlier.

\emph{t}-SNE is a well-known nonlinear DR method which aims to preserve neighborhoods of a given point. Its popularity is arguably due to its good ability to separate similar data clusters present in high-dimensional spaces to create a strong visual separation of clusters in the 2D projection\,\cite{tsne:original}. Recently, Bernard \emph{et al.}\ showed that \emph{t}-SNE is the preferred DR method for labeling, when compared with other methods such as non-metric MDS, Sammon mapping, and Principal Component Analysis (PCA), due to its clear cluster separation\,\cite{labeling_wt_DR2017}. Lewis \emph{et al.}\ also confirmed that users prefer visualization methods that clearly separate clusters (assuming, of course, such clusters exist in the data), as this is seen as a sign of quality of the method\,\cite{dr:lewis2012behavioral,labeling_wt_DR2017}.

However, \emph{t}-SNE's complexity is quadratic in the number of points\,\cite{acctsne}. While accelerated variants exist\,\cite{a_sne,h_sne,gpu_tsne}, these are quite complex to implement and not yet widespread. Additionally, due to its stochastic nature and underlying cost minimization process, it is hard for users to predict the results of \emph{t}-SNE for a given data set and parameter settings\,\cite{tsne:wattenberg}. A more recent competitor, UMAP, has been introduced and used in astronomical applications\,\cite{umap2018, umap:astro}, but to date has not been widely applied and assessed by domain experts in that field.

\subsection{DR in astronomy}
\label{sec:relatedwork:astro}
While applications of DR include biomedicine, computer security, and various other fields\,\cite{tsne:bioinformatics,tsne:security}, our main application domain in this paper is astronomy. We cover next the important use-cases of DR in astronomy to explain the importance, recent works, and limitations of DR, and to elicit the needs of domain experts when using such methods.

Astronomical data sets have long been considered ‘big’ data, and are still growing larger due to the advancement of sensor technology and signal processing capacity. The recent Gaia DR2 catalogue \cite{astro:GAIADR2_1, astro:GAIADR2_2, astro:GALAHDR2} contains more than 1.6 billion objects with tens of dimensions. Sifting through these big data catalogues is an excellent test for DR methods.

High-dimensional data analysis using DR for clustering purposes has widely been used in astronomy, starting with one of the earliest DR methods, PCA\,\cite{astro:clustering_histo_pca}. Since its first application in astronomy\,\cite{Deeming64}, PCA continues to be widely used by astronomers, \emph{e.g.,}\ to explore the space of stellar elemental abundances and describe how many controlling parameters exist\,\cite{pca:astro4_ting}, and to find clusters in that space\,\cite{pca:astro2}.

More recent studies show the use of \emph{t}-SNE in astronomy\,\cite{tsne:astro, tsne:astro2}. Anders \emph{et al.}\,\cite{tsne:astro} show how domain experts use \emph{t}-SNE to manually label interesting points and clusters in the 2D ``abundance-space'' (the term used in astronomy to denote projection space) using a number of stellar abundances as input. However, they attempt to manually label data clusters based on nearby, often-overlapping, and sometimes very small clusters in the 2D projection, which can lead to highly uncertain labels. Moreover, the labels generated did not arise solely from the \emph{t}-SNE projection but also from analysis of a scatter plot matrix of the original abundance-space data in an iterative process with the \emph{t}-SNE projection (see Figure~1 in \cite{tsne:astro}). We explore the above challenges of using \emph{t}-SNE for data labeling in Section~\nameref{sec:astro}.

\par In summary, previous work has shown (1) the importance of DR in data labeling; (2) that a clear separation of clusters provide an intuitive labeling experience by end-users; (3) that \emph{t}-SNE is the current state-of-the-art DR method used in manual labeling and user-guided label propagation; and (4) that \emph{t}-SNE may not always give clear cluster separation, which explains the quest for an alternative DR method that provides a clear visual separation of clusters for users to easily label these clusters.


\section{Proposed Method}
\label{sec:method}
We next present our method, which consists of two main steps: local gradient clustering (Section~\nameref{sec:method:lgc}) and actual projections using RP, LMDS, \emph{t}-SNE, and UMAP (Section~\nameref{sec:sub:dr}).

\subsection{Local Gradient Clustering}
\label{sec:method:lgc}
As outlined in Section~\nameref{sec:relatedWork}, we aim to obtain a high visual cluster separation in a projection by `preconditioning' the DR method. Good candidates for this preconditioning are mean shift-based methods\,\cite{gc1975,gc1995,ms,ms2007,alex1}. Such methods have previously been suggested to estimate modes of clusters, in combination with DR, to cluster and visualize high-dimensional data\,\cite{realworld:banknote_sep}. In contrast, our method does not need the mode information to create a high visual cluster separation.
 
 Mean shift-based methods estimate the sample density using kernel density estimation (KDE)\,\cite{kde1} and iteratively shift samples upwards in the density gradient. For a data set $D=\{\mathbf{x}_i\}$, we define the multivariate kernel density estimator at location $\mathbf{x} \in \mathbb{R}^n$ by
\begin{equation}
\label{Eq1}
	\rho(\mathbf{x}) = \sum_{\mathbf{x}_i \in N(\mathbf{x})} K\left(\frac{\|\mathbf{x}-\mathbf{x}_i\|}{h}\right),
\end{equation}
where $K(\cdot) : \mathbb{R}^{+} \rightarrow \mathbb{R}^{+}$ is a radially-symmetric univariate kernel of bandwidth $h$. $N(\mathbf{x})$ denotes the set of samples $\mathbf{x}_i$ which affect the density $\rho$ at location $\mathbf{x}$. In classical KDE\,\cite{gc1975}, $N(\mathbf{x})=D$, \emph{i.e.}, all samples affect all density locations. Another possibility is to use only samples closer to $\mathbf{x}$ than $h$, \emph{i.e.}, $N(\mathbf{x}) = \{\mathbf{x}_i \in D\,:\,\| \mathbf{x} - \mathbf{x}_i \| < h \}$. This offers a better local control of the scale of patterns (clusters) formed by mean shift and also significantly accelerates the density estimation\,\cite{alex1}. However, this assumes that all data clusters have comparable \emph{scale} in $D$, and that this scale ($h$) is known, which typically is not the case with high-dimensional data sets of varying density. We refine the above model by locally setting $h$ to the distance between $\mathbf{x}$ and its $k_s$-nearest neighbor in $D$. The free parameter $k_s$ thus determines the simplification scale of the data set. More intuitively, all $k_s$-nearest neighbors of a point $\mathbf{x}$ are considered to be in the same cluster as $\mathbf{x}$.

After estimating $\rho$ over $D$, we shift all samples $\mathbf{x}_i \in D$ for $T$ iterations along the density gradient by the following update rule
\begin{equation}
\label{Eq2}
	\mathbf{x}_i^{next} = \mathbf{x}_i + \alpha\frac{\nabla \rho(\mathbf{x}_i)}{\mbox{max}(\| \nabla\rho(\mathbf{x}_i)\|, \epsilon )},
\end{equation}
%
%
\begin{figure}[htb]
  \centering
  \includegraphics[width=0.9\linewidth]{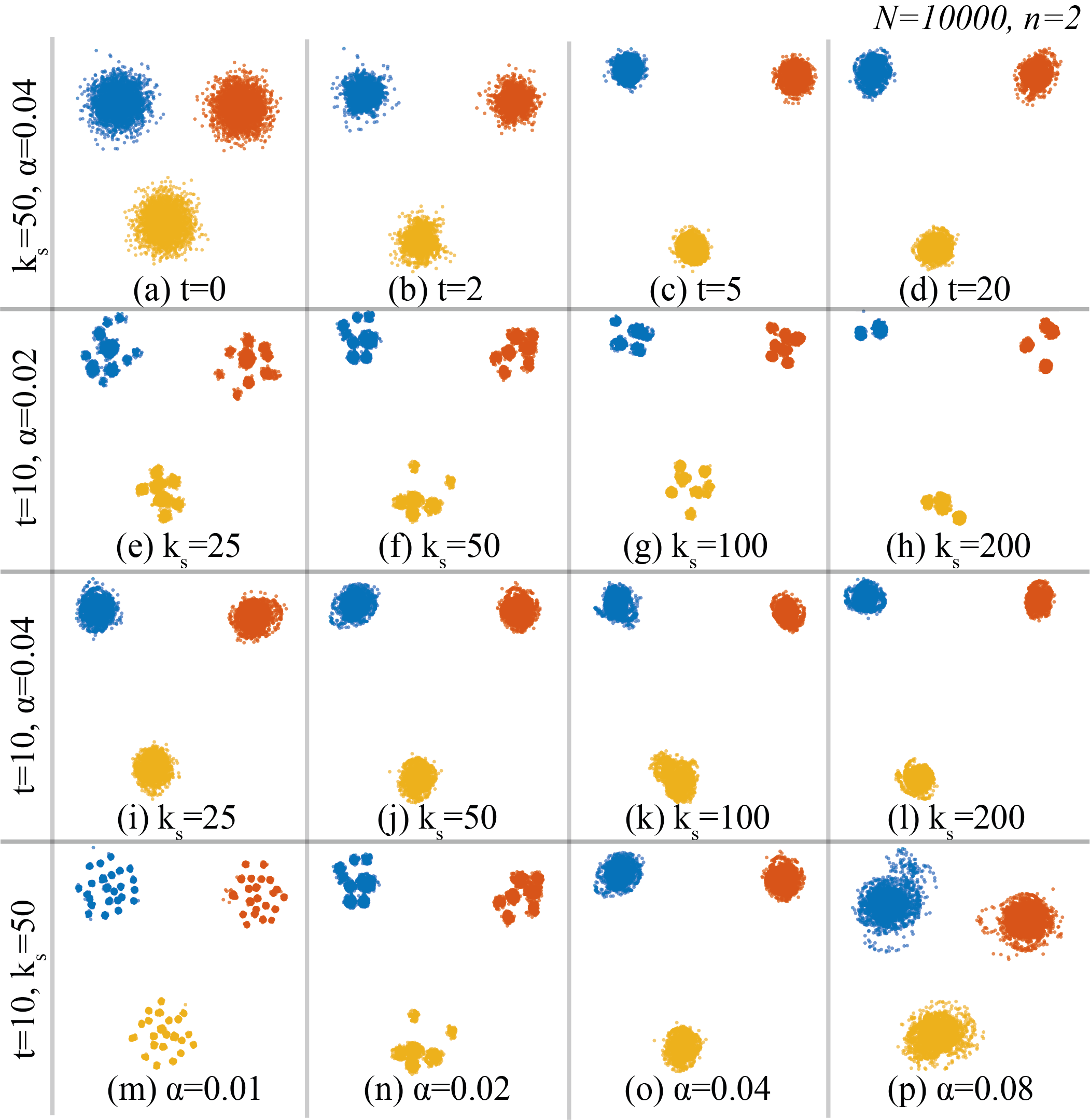}
  \parbox[t]{1\columnwidth}{\relax}
  \caption{\label{fig:1params_gaussian} Effects of parameters used in LGC. 2D Gaussian data with 10$\mathrm{K}$ observations and three clusters (a) are used to show the effects of the number of iterations ($T$) as shown in (a)--(d), number of nearest neighbors ($k_s$) in (e)--(l), and learning rate ($\alpha$) in (m)--(p). Points are color-coded based on their ground-truth labels. The cluster borders become fuzzy when using a too high $T$, as shown in (d). $k_s$ and $\alpha$ both contribute to the degree of segmentation of the clusters; without choosing an appropriate $\alpha$, $k_s$ may not significantly affect the segmentation, as shown in rows (e)--(h) and (i)--(l). Note that $\alpha$ uses a fixed range of $[0,1]$.}
\end{figure}
%
\begin{figure}[htb]
  \centering
  \includegraphics[width=0.9\linewidth]{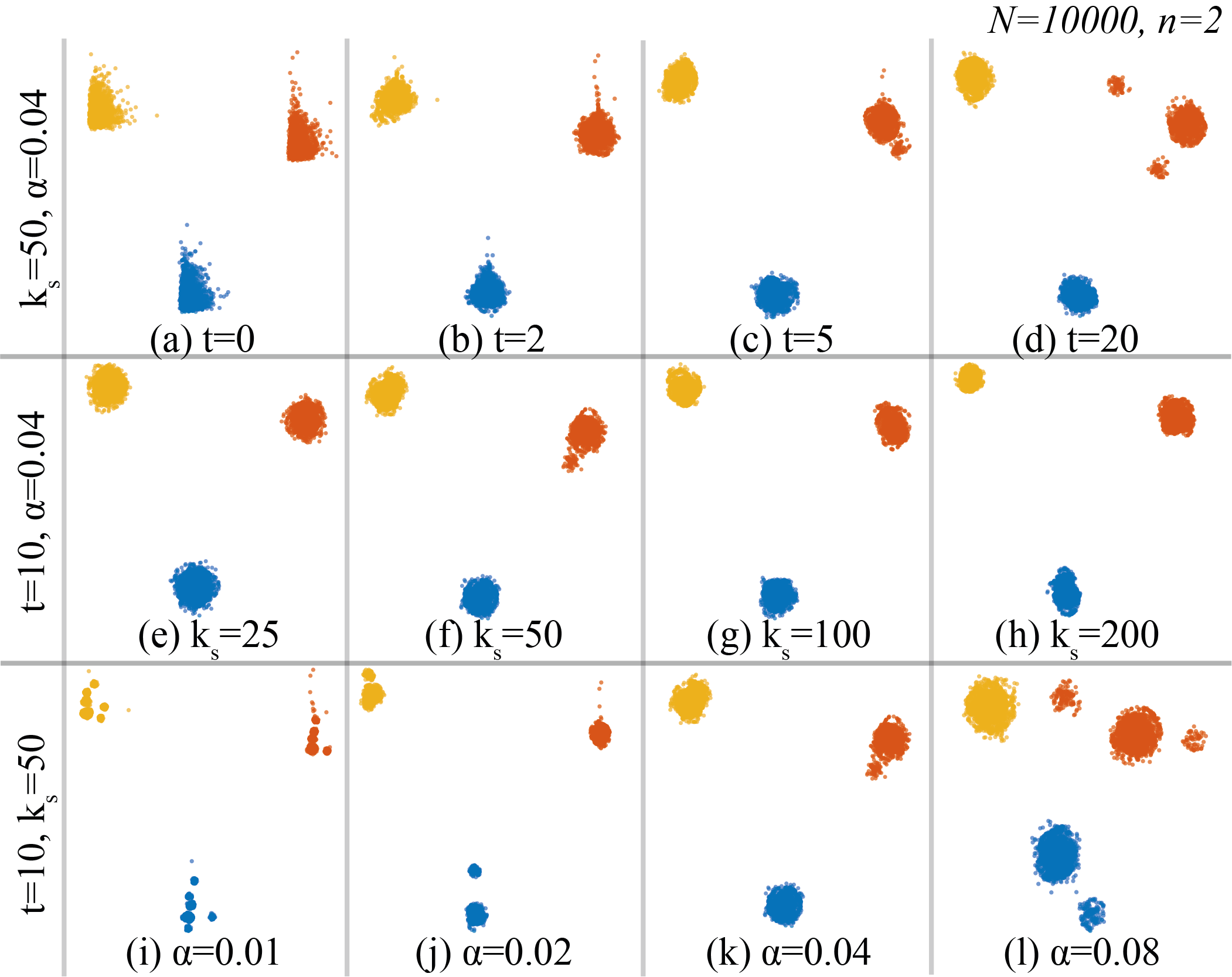}
  \parbox[t]{1\columnwidth}{\relax}
  \caption{\label{fig:2params_non_Gaussian} Effects of different parameters using 2D non-Gaussian (log-normal, $\mu=0$, and $\sigma=1$) data with 10$\mathrm{K}$ observations. The effects of the parameters are similar to those in Figure~\ref{fig:1params_gaussian}. However, LGC with too large values of $T$ and $\alpha$ is prone to outliers (long tails), as shown in (d) and (l). This problem can be solved by setting a larger value of $k_s$.
  }
\end{figure}
where $\alpha$ is the `learning rate', which determines the convergence speed of the process, and $\epsilon=10^{-5}$ is a fixed regularization factor used to handle gradients near zero. For $K$, we use an Epanechnikov (parabolic) kernel, which is optimal for KDE in a mean-squared error sense\,\cite{epanechnikov:mse}. This kernel yields smaller movements (shifts) as compared to a Gaussian kernel, thereby favoring the stability of the process. Note that Eqs.~\ref{Eq1} and~\ref{Eq2} are coupled, as we estimate the gradient $\nabla\rho$ (Eq.~\ref{Eq1}) after every iteration. This means that we perform the nearest neighbor search for every iteration as in Hurter \emph{et al.}\,\cite{alex1}.
In contrast, classical Gradient Clustering (GC)\,\cite{gc1975} performs nearest neighbor search only for the first iteration, and uses those neighbors in subsequent iterations. Hurter \emph{et al.} showed advantages of nearest neighbor search at every iteration in terms of robustness of the sample shift with respect to parameter tuning. Hence, we follow the same approach.
Due to our usage of nearest neighbors, we call our sharpening approach Local Gradient Clustering (LGC), by analogy with Gradient Clustering (GC). The key added value of LGC discussed in this paper is its preconditioning of the data that leads to better results of DR techniques.

Figures~\ref{fig:1params_gaussian}--\ref{fig:2params_non_Gaussian} show the effect of the free parameters $T$ (number of iterations), $k_s$ (number of nearest neighbors), and $\alpha$ (learning rate) for LGC. Color encodes ground-truth labels, which are known for these data sets. Each row of Figures~\ref{fig:1params_gaussian}--\ref{fig:2params_non_Gaussian} shows the results of varying a single parameter, with the other two parameters fixed. The data set $D$ contains synthetic Gaussian random data ($N=10\mathrm{K}$ and $n=2$) for Figure~\ref{fig:1params_gaussian} and non-Gaussian (log-normal, $\mu=0$, and $\sigma=1$) random data ($N=10\mathrm{K}$, $n=2$) for Figure~\ref{fig:2params_non_Gaussian}. We use $n=2$ to demonstrate the effect of each parameter. Indeed, for $n=2$, we can directly look at $D$ to assess LGC without DR. Note that similar behaviors are shown for higher $n$-values. The effect of the three parameters is as follows.

\noindent\textbf{Learning rate $\alpha$:} Controls the \emph{speed} of shifts and affects the degree of segmentation, see the bottom rows of Figure~\ref{fig:1params_gaussian} and \ref{fig:2params_non_Gaussian}. If $\alpha$ is too large, points move too far and can overshoot the mode of a cluster during LGC as shown in Figure~\ref{fig:1params_gaussian}(p) (see also Section~IV-A in\,\cite{gc1975}). Conversely, too small $\alpha$-values yield too small shifts (Figure~\ref{fig:1params_gaussian}(m)) and thus can result in an oversegmentation of the data (too many small clusters). The interconnection between $\alpha$ and $k_s$ is discussed further in Section~\nameref{sec:discussion:parameters}.

\noindent\textbf{Nearest neighbors $k_s$:} Controls how \emph{localized} a shift is. Both $k_s$ and $\alpha$ affect the degree of segmentation; yet, without choosing an appropriate $\alpha$, $k_s$ may not significantly affect the segmentation, as shown in the second and third rows of Figure~\ref{fig:1params_gaussian}. Here, we empirically fix $k_s=50$ based on the stability and speed of our method. Too small $k_s$-values can create oversegmentation (many small clusters) and can sharpen dense areas of noise making our method unstable (see detailed discussion in Section~\nameref{sec:discussion:sharpenNoise}); a too large value of $k_s$ increases the number of nearest neighbor searches resulting in slower computation (see Section~\nameref{sec:sub:scalability:speed}).

\noindent\textbf{Number of iterations $T$:} This parameter controls the amount of cluster \emph{separation}. If $T$ is too small, points will shift only a few steps along the density gradient, resulting in little difference from the original data. We have observed that intra-cluster points are close enough for clusters to be visually well separated using $T=5$ for Gaussian synthetic data and $T=10$ for non-Gaussian synthetic data. Varying $T=10$ by a factor of two may not significantly change the obtained result, but too many iterations also add to the computing time (discussed next in Section~\nameref{sec:sub:scalability:speed}). Setting $T=10$ for all experiments in this paper allows us to obtain a data separation that is sufficient to yield a clear visual separation in the DR projection of the preprocessed data. 

Points can overshoot the local mode given their $k_s$-nearest neighbors when using a too-large $T$-value. This is why the borders of clusters in Figure~\ref{fig:1params_gaussian}--\ref{fig:2params_non_Gaussian}(d) become fuzzier compared to those in Figure~\ref{fig:1params_gaussian}--\ref{fig:2params_non_Gaussian}(c). This can be solved by using a smaller value of $\alpha$. A similar issue is solved by decreasing the advection step in time\,\cite{alex1}. However, in that context, the aim was to collapse close data points to a \emph{single} point. This is not the aim of VCS, so we cannot use that approach in our context.

\par Summarizing, we can use a single free parameter $\alpha$ to control the sharpening step after fixing the values of $k_s$ and $T$. The effects of $\alpha$ on speed are discussed separately in Section~\nameref{sec:sub:scalability:speed}. We implemented LGC in C++ for higher-speed performance, using Nanoflann\,\cite{flann_2009,flann_2014,nanoflann} for the nearest neighbor search in $\mathbb{R}^n$. We have also evaluated other nearest-neighbor search algorithms (see Section~\nameref{sec:sub:scalability:speed}). Our code is publicly available\,\cite{hd-sdr}.

\subsection{Dimensionality Reduction Candidates for HD-SDR}
\label{sec:sub:dr}
\par As explained in Section 1, the aim of our method is to improve the visual cluster separation for existing DR methods which are lacking in this respect; and do this in a computationally efficient way and with minimal parameter-setting effort. We have achieved the first concern (cluster separation) in the data space by using LGC (Section~\nameref{sec:method:lgc}). Now we test our method on several DR techniques that take the LGC-sharpened data as input and project it. We use three different DR methods from the publicly-available C++ Tapkee toolkit\,\cite{tapkee:cite1}, as well as UMAP available in Python. These are selected based on the following requirements:
\begin{itemize}
    \item no prior knowledge (labels) of the data;
    \item computational scalability to large data sets (tens of thousands of samples, hundreds of dimensions);
    \item ease of use in terms of free parameters with documented presets;
    \item showing (weak or strong) visual cluster separation.
\end{itemize}

Adding to the last requirement, our aim is to sharpen clusters in $n$D so that the clusters are also visually separable after DR, rather than creating clusters via DR methods that show no clustering ability. This is why we select DR methods that exhibit different degrees of cluster separation. Random Projection (RP), Landmark Multidimensional Scaling (LMDS), \emph{t}-SNE, and UMAP are the methods that best meet the above criteria\,\cite{RP1:randomProj,lmds:original,tsne:original,umap2018}. The quantitative survey of DR methods of Espadato \emph{et al.}\,\cite{mateusDR_survey2019} found \emph{t}-SNE, UMAP, Projection By Clustering (PBC), and Interactive Document Maps (IDMAP) to have the highest global quality. Here, we use UMAP because it is a strong competitor of \emph{t}-SNE and has also been recently applied to astronomy\,\cite{umap:astro}, our main application domain. Empirically, UMAP, \emph{t}-SNE, LMDS, and RP, in descending order, show the strongest cluster separation in our study. Apart from the above, note that any DR method can be used in our proposed approach. To show this, we apply our sharpening method on a labeled real-world WiFi data set and feed it to the DR implementations from\,\cite{mateusDR_survey2019} (see supplemental material and Section~\nameref{limitations}).

We briefly introduce the selected DR techniques used in this paper. Note that \emph{t}-SNE was already explained in Section~\nameref{sec:relatedWork}.

\noindent\textbf{Random Projection (RP):} Although nonlinear projections achieve better distance preservation for high-dimensional data\,\cite{sorzano14,isomap:original}, we use the linear DR technique Random Projection (RP) to demonstrate the sharpening effect on a DR with relatively poor cluster separation. RP projects a random matrix consisting of orthogonal unit vectors to lower dimensions, aiming to preserve pairwise distances. RP needs less memory and is faster to compute than PCA\,\cite{RP1:randomProj,rp2:randomproj_survey,pca}. RP is of order $O(N\times n\times s)$, where the $N$ samples in $\mathbb{R}^n$ are projected to an $s$-dimensional subspace\,\cite{RP1:randomProj,rp2:randomproj_survey}.

\noindent\textbf{Landmark Multidimensional Scaling (LMDS)} is a nonlinear variant of Multidimensional Scaling\,\cite{MDS}. Computational scalability (linear in the sample count $N$) is achieved by projecting a small subset of the so-called landmark samples (5\% of $N$ in most of our experiments) by classical MDS, around which remaining samples are projected using a fast triangulation procedure\,\cite{lmds:original}. For completeness, we mention that we also used LMDS with increasingly more landmarks and obtained visually similar results (not included here for brevity).

\noindent\textbf{Uniform Manifold Approximation and Projection (UMAP)} is a recent competitor to \emph{t}-SNE due to its strong separability of clusters. UMAP's model assumes that the data is close to being uniformly distributed on a Riemannian manifold; the Riemannian metric is locally constant; and the manifold is locally connected\,\cite{umap2018}. UMAP aims to find the lower, \emph{i.e.,} two-dimensional embedding with a topological structure that best represents the fuzzy topological structure of the original data\,\cite{umap2018}.

\section{Results}
\label{sec:results}
We compare HD-SDR with DR on both synthetic data (Section~\nameref{sec:results:synthetic}) and real-world data (Section~\nameref{sec:results:realworld}). We run all our experiments on a PC having a Core i7-8650U (2.11GHz) processor with 16G RAM.

\begin{sidewaysfigure*}
  \includegraphics{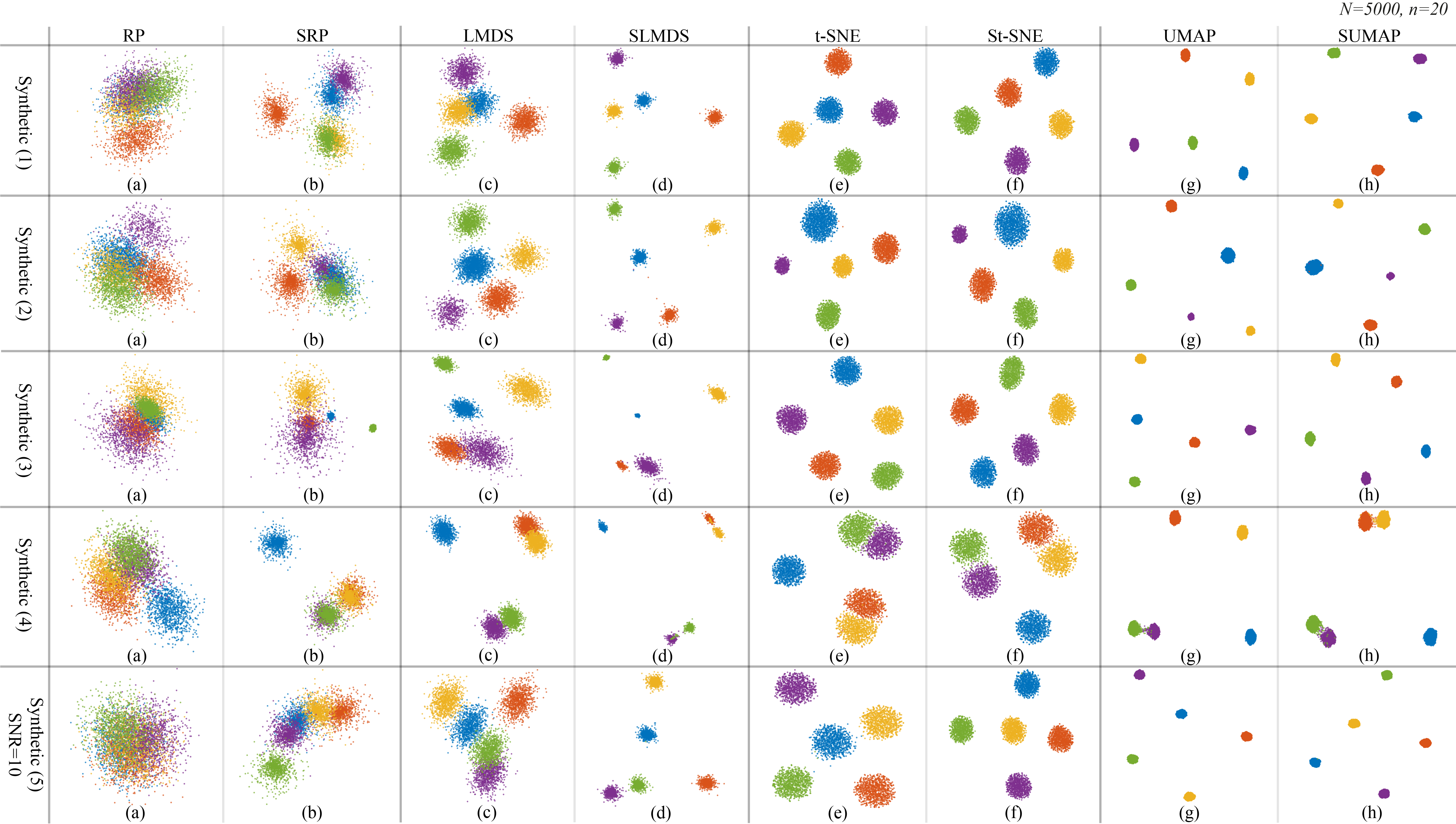}
  \caption{\label{fig:3fdr_all_synthetic} Comparison of DR and HD-SDR using five different types of synthetic data. DR methods are ordered from left to right based on how well clusters are separated. The samples are colored by the ground-truth labels of five different clusters for visual examination purposes. The results for SRP and SLMDS have been obtained with $\alpha=0.04$ and for S\emph{t}-SNE (perplexity$=50$) and SUMAP with $\alpha=0.01$. Note that our sharpening method enhances the separation of clusters for DR methods with less ability to separate clusters (i.e., RP and LMDS). The HD-SDR is however less effective for separating subclusters, as shown in the fourth row.
  }
\end{sidewaysfigure*}

\begin{sidewaysfigure*}
  \centering
  \includegraphics{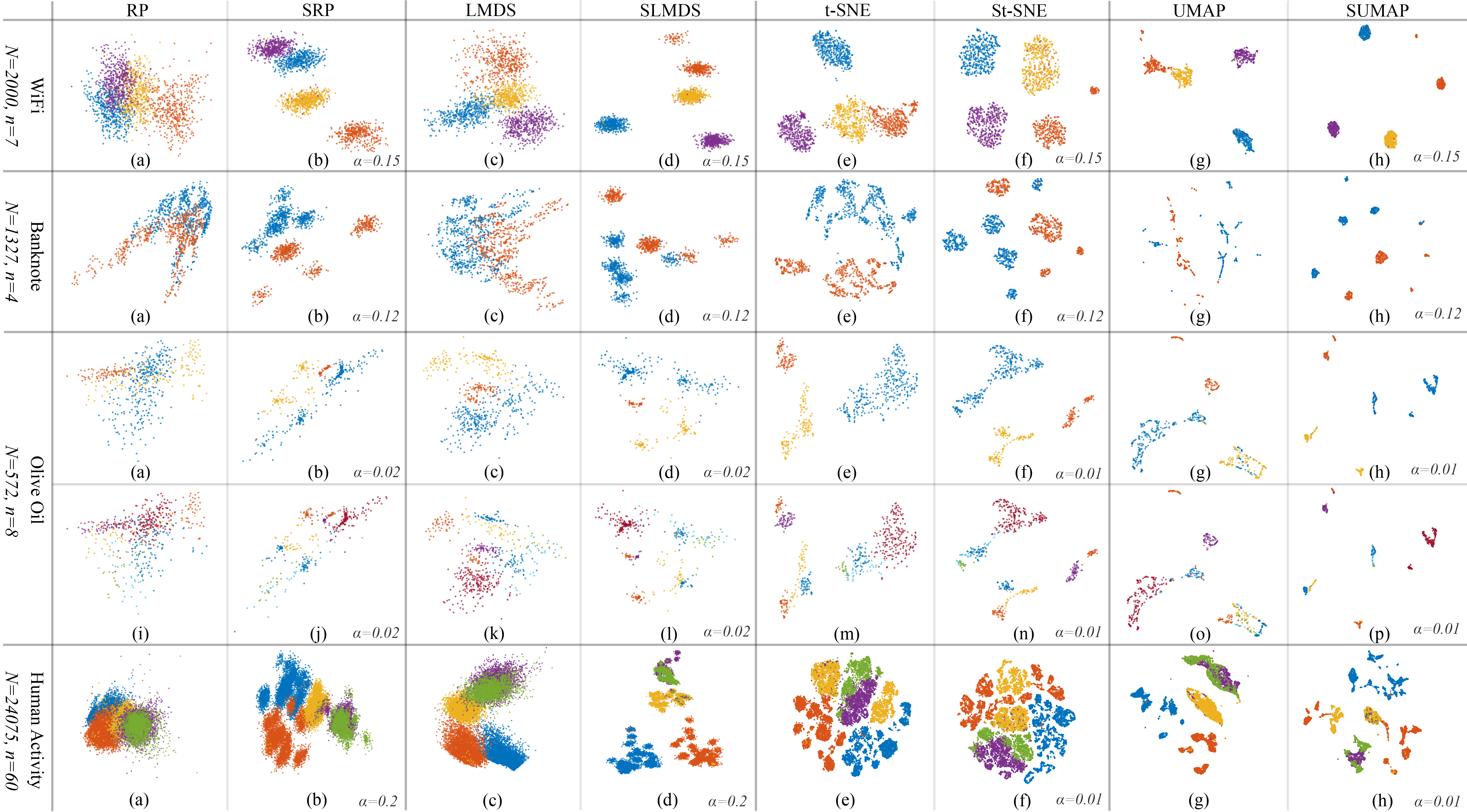}
  \caption{\label{fig:4fdr_all_realworld} Comparison of DR and HD-SDR using four different real-world data. DR methods are ordered from left to right based on how well clusters are separated. The samples are colored by their ground-truth labels for visual examination purposes. Note that the sharpening method significantly enhances the visual separability of clusters for the first three data sets, except for SUMAP as shown in the eighth column. Additionally, S\emph{t}-SNE and SUMAP exhibit more oversegmentation of clusters than SRP and SLMDS.
  }
\end{sidewaysfigure*}

\subsection{Synthetic data: qualitative evaluation}
\label{sec:results:synthetic}
We generated Gaussian random data consisting of five clusters ($N=5\mathrm{K}$, $n=20$) to cover five types of inter-sample distance distributions:
\begin{itemize}
    \item[(1)] even spread of inter-cluster distances with equal intra-cluster densities with equal Gaussian variance;
    \item[(2)] even spread of inter-cluster distances with different intra-cluster densities;
    \item[(3)] uneven spread of inter-cluster distances (skewed distribution);
    \item[(4)] two pairs of subclusters and a single cluster;
    \item[(5)] noise added to (1) with a signal-to-noise ratio (SNR) of 10.
\end{itemize}

This way, we can explicitly control the clusters and their separation in the data, and thus assess how well the 2D projections capture this separation. We randomly generate five trials per data set type above and show the results of a single trial in Figure~\ref{fig:3fdr_all_synthetic}. The five trials are later used for a quantitative evaluation in Section~\nameref{sec:qualityEval}. For synthetic data type (5), we add Gaussian noise using the standard deviation ($\sigma$) calculated by the definition $\mbox{SNR}=10 log_{10}{P_s/\sigma^2} (dB)$, where $P_s$ is the power of the signal.

In Figure~\ref{fig:3fdr_all_synthetic}, the five synthetic data set types (one per row) are projected using both the sharpened and unsharpened versions of RP, LMDS, \emph{t}-SNE, and UMAP. The sharpened versions are denoted by the prefix `S', \emph{i.e.}, SRP, SLMDS, S\emph{t}-SNE, and SUMAP. DR methods are ordered from left to right based on how well they separate clusters. Samples are colored by the cluster labels for visual examination. Here, the LGC parameter $\alpha$ is found empirically by searching the fixed range $[0,1]$ following the explanations in Section~\nameref{sec:method:lgc}. This led us to using $\alpha=0.04$ for SRP and SLMDS and $\alpha=0.01$ for S\emph{t}-SNE (perplexity$=50$) and SUMAP.

Figure~\ref{fig:3fdr_all_synthetic} (first four columns) shows that SRP and SLMDS significantly reduce the amount of overlap between clusters in RP and LMDS for all data sets. For \emph{t}-SNE and UMAP, LGC does not improve cluster separation (Figure~\ref{fig:3fdr_all_synthetic}, last four columns). This is expected, since \emph{t}-SNE and UMAP \emph{already} have a good cluster separation, while RP and LMDS do not. We also see that LGC performs worst for synthetic data which consists of subclusters (4), although SRP and SLMDS show some small improvements in visual cluster separation. In the worst case (Figure~\ref{fig:3fdr_all_synthetic}(g)), SUMAP performs worse than UMAP in separating subclusters using even a small $\alpha=0.01$. Finally, in the fifth row of Figure~\ref{fig:3fdr_all_synthetic}, SRP and SLMDS show slightly better cluster separations of noise-added data compared with RP and LMDS, respectively. For the same data set, St-SNE and SUMAP show a similar cluster separation as their counterparts, t-SNE and UMAP.


\subsection{Real-world data: qualitative evaluation}
\label{sec:results:realworld}
Our method can be applied to any type of tabular data. We next compare HD-SDR and DR using a collection of real-world data of different kinds of data traits.

\begin{table*}[!htbp]
\tiny
\centering
 \caption{Trait values for real-world data sets}
 \label{table:traits}
\resizebox{\textwidth}{!}{%
\begin{tabular}{cccccc}
\hline

Data sets     & Size ($N$) & Dimensionality ($n$) & IDR ($\upsilon$) & Classes ($g$) & Subclasses ($g_{sub}$) \\ \hline

WiFi  & medium (2000)      & low (7)          & high (0.6667)   & medium (4)  & -    \\ 
Banknote   & medium (1327)        & low (4)        & high (0.5)    & small (2)   & - \\ 
Olive Oil       & small (572)       & low (8)          & medium (0.1250)  & medium (3)  & large (9) \\ 
HAD        & large (24075)       & medium (60)          & low (0.0167)   & medium (5) & -    \\ \hline
\end{tabular}
}
\end{table*}

\subsubsection{Data sets and their traits}
\label{sec:data_traits}
We characterize data sets using the traits discussed in Espadoto \emph{et al.}\,\cite{mateusDR_survey2019}. We exclude the \emph{Type} and \emph{Sparsity ratio} traits since we focus here only on dense tabular data. We add the \emph{Classes} trait
that describes the number of clusters the data consists of.

\noindent\textbf{Size} $N$: The number of samples, having three ranges: \emph{small} ($N \leq 1000$); \emph{medium} ($1000 < N \leq 3000$); and \emph{large} ($N > 3000$).

\noindent\textbf{Dimensionality} $n$: The number of dimensions, having three ranges: \emph{low} ($n < 10$); \emph{medium} ($10 \leq n < 100$); and \emph{high} ($n \geq 100$).

\noindent\textbf{Intrinsic dimensionality ratio (IDR)} $\upsilon$: The fraction of the $n$ principal components needed to explain 95\% of the data variance. We use three ranges: \emph{low} ($\upsilon < 0.1$); \emph{medium} ($0.1 \leq \upsilon < 0.5$); and \emph{high} ($0.5 \leq \upsilon \leq 1$).

\noindent\textbf{Classes} $g$: The number of classes (ground-truth labels), having three ranges: \emph{small} ($g \leq 2$); \emph{medium} ($2 < g \leq 5$); and \emph{large} ($g > 5$). We separately measure if the data has sub-classes and count these in $g_{sub}$. Note that we use labels as the ground-truth because there is no other ground-truth to define meaningful clusters for the concrete data sets in our paper.


Table~\ref{table:traits} shows the data sets used for evaluation and their traits.

\noindent\textbf{Banknote}: This data set has $n=4$ features extracted using the Fast Wavelet Transform from $N=1327$ gray scale images of banknotes\,\cite{realworld:uciML}. Each sample (banknote) is labeled as genuine or forged, and the data set is used to train classifiers to predict this label\,\cite{bank_classif}. Projections are used to assess classification: If they show clearly separated different-label clusters, then the features can very likely discriminate between the labels\,\cite{rauber}. We know this data set is easy to classify with accuracy close to 95\%\,\cite{rak,bank_classif}, so our projections should show well-separated clusters.

\noindent\textbf{WiFi}: This data set consists of WiFi signal strengths from various routers measured at four indoor locations\,\cite{realworld:uciML,realworld:wifi,realworld:wifi2}. The data set has $N=2000$ samples with $n=7$ dimensions and is known to have four well-separated clusters\,\cite{realworld:wifi_clustering}. 

\noindent\textbf{Olive oil}: The data has $N=572$ samples of olive oil with $n=8$ dimensions (fatty acid concentrations), with ground-truth label denoting one of the locations in Italy from where the oil was collected. The location consists of three super-classes (North, South, and the island of Sardinia) and sub-classes (three from the North, four from the South, and two from Sardinia).

\noindent\textbf{Human Activity Data (HAD)}: This data set consists of $N=24075$ samples of accelerometer data of a smartphone, each with $n=60$ dimensions\,\cite{realworld:humanActivity,realworld:humanActivity2:calibration,realworld:humanActivity3:feature}. The data is used to classify five motion-related human activities (sit, stand, walk, run, and dance).

\subsubsection{HD-SDR applied to real-world data sets}
Figure~\ref{fig:4fdr_all_realworld} shows the results of HD-SDR applied to our real-world data. DR methods are ordered from left to right based on how well clusters are separated. The parameter settings for HD-SDR and \emph{t}-SNE are displayed together with the plots in Figure~\ref{fig:4fdr_all_realworld}. Overall, HD-SDR yield clearer visual cluster separations than those of the corresponding DR methods without LGC. 

The effect of LGC is more prominent when the underlying DR shows poor cluster separation, such as RP (a) and LMDS (c). Compared to these, SRP (b) and SLMDS (d) show a clear improvement. For the Banknote data set, sharpening significantly reduces overlaps between the two classes (genuine and forged) for RP and LMDS, which are DR methods that show poor cluster separation. A similar difference is visible for the HAD data set (Figure~\ref{fig:4fdr_all_realworld}, last row). For DR methods that exhibit a strong cluster separation, i.e. {t}-SNE and UMAP, the sharpening method improves the visual separation of clusters only by a small degree. 
Furthermore, S\emph{t}-SNE in (f) for the Banknote data set shows more visual subclusters than \emph{t}-SNE (c). Note that all the projections for the Banknote data set exhibit serveral subclusters, which are also found in recent work (compare Figure~\ref{fig:4fdr_all_realworld} second row with Figure 12(a) in \cite{realworld:banknote_sep}). For the Olive Oil data set, the subclusters are already known (see Section~\nameref{sec:data_traits}), 
which is why we color code its projections using both class and sub-class labels (Figure~\ref{fig:4fdr_all_realworld}, third and fourth rows, respectively). We see that sub-classes are revealed by our projections, but not as well as the classes, similar to our experiments on synthetic data (Section~\nameref{sec:results:synthetic}, synthetic data type (4)).

We also see an oversegmentation in HAD data projections. Oversegmentation is worse for S\emph{t}-SNE and SUMAP than SRP and SLMDS, as shown in (e)--(h). This can be solved by using a larger $\alpha$ or by changing $k_s$ (explained in Section~\nameref{sec:method:lgc}). Further discussion of over- and under-segmentation is given in Section~\nameref{sec:discussion}.

In summary, LGC significantly enhances the visual separation of clusters for the WiFi, Banknote, and Olive Oil data, except for SUMAP, where it does not greatly improve upon UMAP (Figure~\ref{fig:4fdr_all_realworld}, 1st column). On the other hand, S\emph{t}-SNE and SUMAP show more oversegmentation than SRP and SLMDS, which suggests that LGC amplifies oversegmentation existing in a base projection.

\begin{figure*}[htb]
  \centering
  \includegraphics[width=0.9\linewidth]{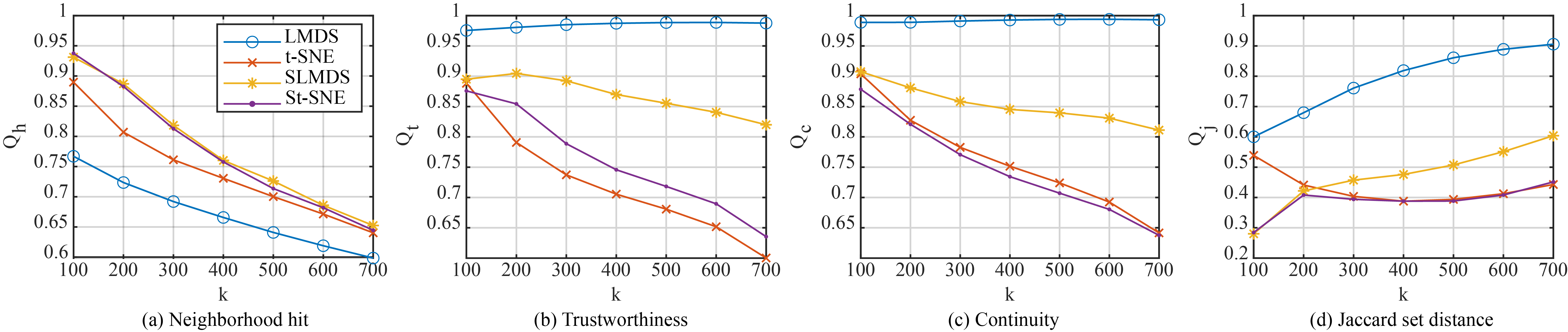}
  \parbox[t]{1\columnwidth}{\relax}
  \caption{\label{fig:5qm_banknote} Results of four neighborhood-based quality metrics for Banknote data: Neighborhood-hit ($Q_h$), Trustworthiness ($Q_t$), Continuity ($Q_c$), and Jaccard set distance ($Q_j$). Note that $Q_h$ is consistent with results from Figure~\ref{fig:3fdr_all_synthetic} and best represents the visual cluster separation, whereas $Q_t$, $Q_c$, and $Q_j$ suggest the opposite. Note that $Q_t$, $Q_c$, and $Q_j$ do not consider class label information. More results including $Q_t$, $Q_c$, and $Q_j$ for the five synthetic data sets can be found in the supplemental materials.}
\end{figure*}

\begin{figure}[htb]
  \centering
  \includegraphics[width=0.9\linewidth]{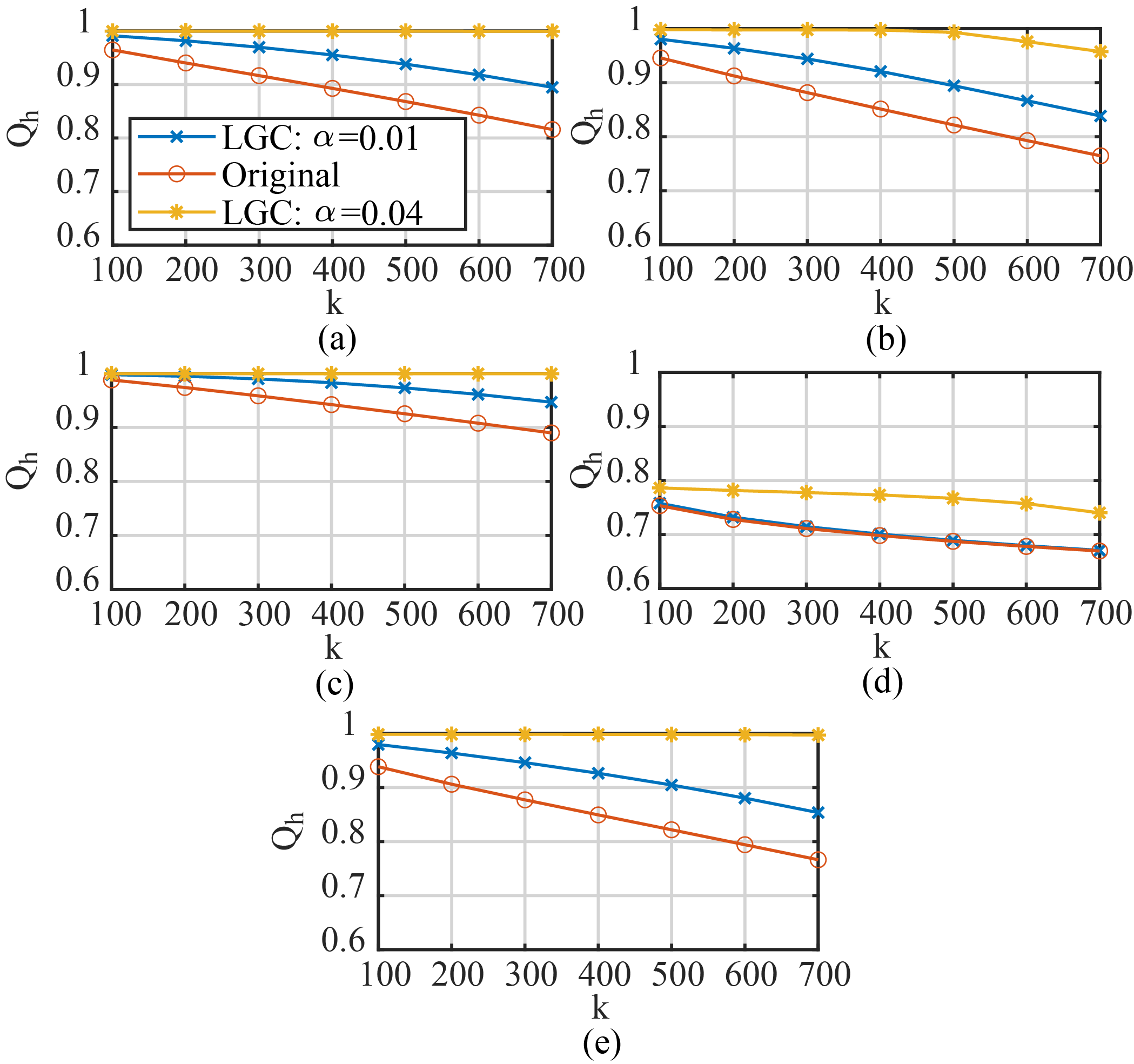}
  \parbox[t]{1\columnwidth}{\relax}
  \caption{\label{fig:6qh_nd_synthetic} Comparison of neighborhood-hit ($Q_h$) for sharpened data and original data of the five different types of synthetic data used in Figure~\ref{fig:3fdr_all_synthetic}. For all synthetic data sets, $Q_h$ is always higher for the sharpened data as compared with the original data. We also note that $Q_h$ for sharpened data is higher when clusters are more separated ($\alpha=0.04$ compared with $\alpha=0.01$).
  }
\end{figure}

\begin{figure}[htb]
  \centering
  \includegraphics[width=0.9\linewidth]{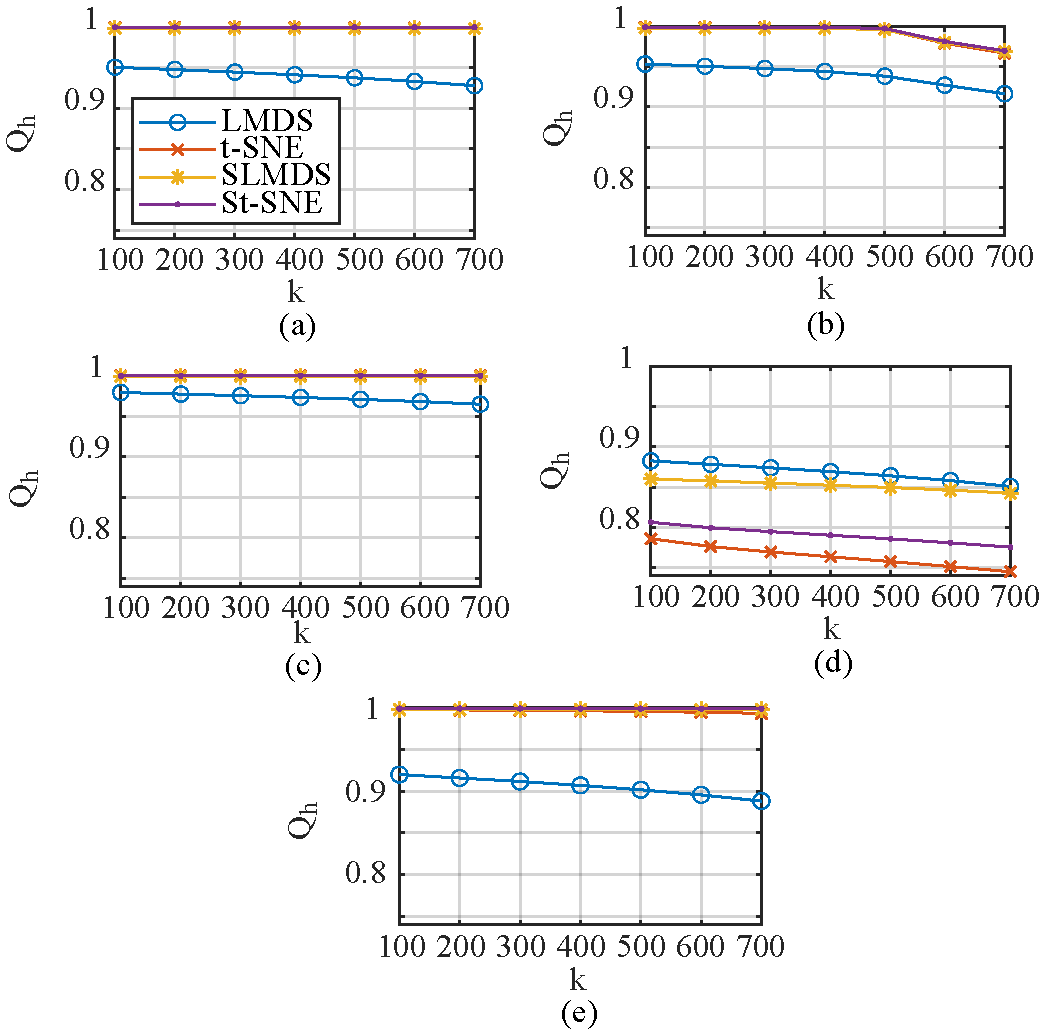}
  \parbox[t]{1\columnwidth}{\relax}
  \caption{\label{fig:7qh_2d_synthetic} Comparison of neighborhood-hit ($Q_h$) for DR and HD-SDR of the five different types of synthetic data used for Figure~\ref{fig:3fdr_all_synthetic}. Note that S\emph{t}-SNE, \emph{t}-SNE, and SLMDS yield high $Q_h$-values near one for (a)--(c) and (e), which suggests that the corresponding labels of the $k$-size neighborhoods are well-preserved for HD-SDR. However, HD-SDR for sub-clustered data produces lower $Q_h$ compared with DR, as shown in (d), and this can be seen visually in Figure~\ref{fig:3fdr_all_synthetic}(a)--(d). More results including $Q_t$, $Q_c$, and $Q_j$ for the five synthetic data sets can be found in the supplemental materials.
  }
\end{figure}

\subsection{Quantitative Evaluation}
\label{sec:qualityEval}
%
While there are perception-based evaluations with extensive user studies on projection methods\,\cite{perceptionEvalDR_userstudy1}, we evaluate here the projection methods quantitatively using quality metrics. As explained in Section~\nameref{sec:relatedwork:VCS}, visual cluster separation is an important property of projection methods which we aim to evaluate for our proposed HD-SDR method. To do this, we need the function $H$ to quantify clusters both in the data space $D$ and projection space $P(D)$. There is, however, no \emph{unique} way to measure the presence, extent, or even count of the clusters in such spaces. Hence, we next use `weak forms' of $H$ given by projection quality metrics. These are functions $Q : (D,P(D)) \rightarrow \mathbb{R}^{+}$. High $Q$-values for a projection $P(D)$ indicate that $P(D)$ preserves the data structure of $D$ -- in which case $H(P(D))$ should be close to $H(D)$. In particular, if LGC brings added value, we should see that $Q(LGC(D),P(LGC(D))) \geq Q(D,P(D))$ for various data sets $D$ and projections $P$.

We focus specifically on neighborhood-based metrics, which are better than distance-based metrics when assessing tasks related to finding clusters in the data\,\cite{jaccard_martins2015,mateusDR_survey2019}. From these, we consider the following four metrics.

\noindent\textbf{Trustworthiness ($Q_t$) and continuity ($Q_c$)} relate to errors produced by false
neighbors (points that are neighbors in $P(D)$ but not in $D$) and missing neighbors (points that are neighbors in $D$ but not in $P(D)$), respectively\,\cite{trustworthiness_orig}. Formally put:
\begin{equation}
    \label{Eq3}
    	Q_t(k) = 1-\frac{2}{N k(2N-3 k-1)} \sum_{i=1}^{N} \sum_{j \in U_{k} (i)} (r(i,j)-k),
\end{equation}

\begin{equation}
    \label{Eq4}
    	Q_c(k) = 1-\frac{2}{N k(2N-3 k-1)} \sum_{i=1}^{N} \sum_{j \in V_{k} (i)} (\hat{r}(i,j)-k),
\end{equation}
where $U_k$ and $V_k$ are the set of false neighbors and missing $k$-nearest neighbors of point $i$, respectively; $r(i,j)$ is the rank of point $j$ in the ordered set of neighbors of point $i$ in $D$; and $\hat{r}(i,j)$ refers to the rank of point $j$ in the ordered set of neighbors of point $i$ in $P(D)$. While $Q_t$ measures the credibility of neighborhood relationships in the projection, $Q_c$ captures the discontinuities of the projection caused by missing neighbors\,\cite{trustworthiness_orig}. $Q_t$ and $Q_c$ lie in the range of $0$ (worst) to $1$ (best).

\noindent\textbf{Jaccard set distance ($Q_j$)} measures the fraction of the $k$-nearest neighbors of a point in $P(D)$ that are also among the $k$-nearest neighbors of that point in $D$\,\cite{jaccard_martins2015,jaccard_original}. We average $Q_j$ over all points, leading to 
\begin{equation}
    \label{Eq5}
    	Q_j(k) = \frac{1}{N} \sum_{i=1}^{N} \frac{|W_k^2(i) \cap W_k^n(i)|}{|W_k^2(i) \cup W_k^n(i)|},
\end{equation}
where $W_k^2(i)$ and $W_k^n(i)$ are the sets of the $k$-nearest neighbors of point $i$ in $P(D)$ and $D$, respectively. This metric also lies in the range $[0, 1]$. Low values indicate that neighbors are poorly preserved and conversely for high values. 
    
\begin{figure*}[hbt]
  \centering
  \includegraphics[width=0.9\linewidth]{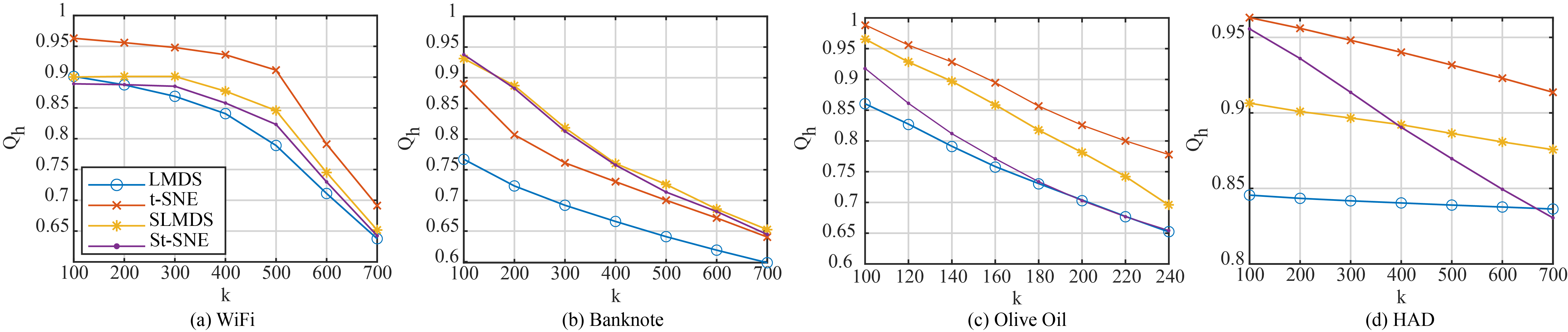}
  \parbox[t]{1\columnwidth}{\relax}
  \caption{\label{fig:8qh_realworld} The neighborhood-hit ($Q_h$) metric for DR and HD-SDR using labeled real-world data sets with different values of $k$. We note that $Q_h$ is lower for S\emph{t}-SNE than for \emph{t}-SNE for most values of $k$ in (b)--(d). However for LMDS, which produces a weaker separation of clusters than \emph{t}-SNE, SLMDS produces a higher value of $Q_h$ compared with LMDS for all data sets.}
\end{figure*}
    
\noindent\textbf{Neighborhood-hit ($Q_h$)} measures the proportion of $k$-nearest neighbors of a given point that fall into the same class (have the same ground-truth labels), averaged over all data points\,\cite{neighborhoodhit,coimbra21}. It ranges between $[0,1]$ and is defined as
\begin{equation}
    \label{Eq6}
    	Q_h(k) = \frac{1}{N} \sum_{i=1}^{N} \frac{|G_k^n(i)|}{k},
\end{equation}
\begin{equation}
    \label{Eq7}
        G_k^n(i)=\{j | g_j = g_i, j \in W_k^2(i)\},
\end{equation}
with $W_k^2(i)$ defined as earlier and $g_i$ being the ground-truth labels (classes) of points $i$. $Q_h$ is often used in classifier evaluation\,\cite{mateusDR_survey2019}.
A discussion on the interpretation of $Q_h$, $Q_j$, $Q_t$ and $Q_c$ is given next in Section~\nameref{sec:evaluation_HD-SDR} when analyzing the values of these metrics for both synthetic and real-world data.

\subsubsection{Evaluation of LGC}
\label{sec:evaluation_LGC}
To capture whether the neighbors and their corresponding labels are preserved well by LGC, we measure $Q_h$ on the sharpened data ($Q_h(LGC(D),P(LGC(D))$) and compare it with $Q_h$ measured on the original data ($Q_h(D,P(D))$). For clear VCS, we expect $Q_h(LGC(D),P(LGC(D)) \approx 1$ and being larger than $Q_h(D,P(D))$.

Figure~\ref{fig:6qh_nd_synthetic} shows the average $Q_h$ over our five data set types for different $k$-values (see Section~\nameref{sec:results:synthetic}). We see high $Q_h$-values for (a)--(c) and (e), suggesting that LGC has achieved the desired sharpening effect. Although the data set (d) shows lower $Q_h$-values than (a)--(c) and (e), the values are still higher than the $Q_h$-value of the original, unsharpened, data.
We also see that $Q_h$ increases for $\alpha=0.04$ (yellow curves) as compared to $\alpha=0.01$ (blue curves). This is in line with Figure~\ref{fig:3fdr_all_synthetic} which also uses $\alpha=0.04$. Lastly, we see how $Q_h$ decreases with $k$. The $Q_h$ decreases is significant for $k >1000$ (not shown in the figure). This is expected, since our synthetic data clusters have 1000 points per cluster.

\subsubsection{Evaluation of HD-SDR}
\label{sec:evaluation_HD-SDR}
We next evaluate $Q_h$ for LMDS, SLMDS, \emph{t}-SNE, and S\emph{t}-SNE. We evaluate LMDS and t-SNE and their SDR results to show the difference between the two methods that have different degrees of cluster separation. A higher $Q_h$ for the HD-SDR methods (SLMDS and S\emph{t}-SNE) indicates that our proposed sharpening yields better VCS than the original DR methods.

\noindent\textbf{Synthetic data:} Figure~\ref{fig:7qh_2d_synthetic} shows $Q_h$ for our five synthetic data types for different $k$-values. For each $k$, we show the average $Q_h$ over all data sets of that type. For cases (a), (b), and (e), S\emph{t}-SNE, \emph{t}-SNE, and SLMDS have the highest $Q_h$-values in order, while LMDS scores lowest. For data set (c), \emph{t}-SNE yields a slightly higher $Q_h$ than S\emph{t}-SNE, but both values are close to one. This is in line with the projections in Figure~\ref{fig:3fdr_all_synthetic} (third row) which show well-separated clusters for both \emph{t}-SNE and S\emph{t}-SNE. For case (d) we can see that, although the visual separation is clearer for SLMDS than for LMDS, the subclusters are mixed using synthetic data type (4) in Figure~\ref{fig:3fdr_all_synthetic}, which is why $Q_h$ is lower for SLMDS than LMDS. On the other hand, S\emph{t}-SNE creates a slightly better separation of subclusters than \emph{t}-SNE in Figure~\ref{fig:3fdr_all_synthetic}, which is why $Q_h$ is higher for S\emph{t}-SNE than for \emph{t}-SNE in Figure~\ref{fig:7qh_2d_synthetic}d. Furthermore, Figure~\ref{fig:7qh_2d_synthetic}(e) shows that our method is noise-resistant up to $\mbox{SNR}=10$. 
Figure 1 (supplemental) shows the corresponding $Q_t$, $Q_c$, and $Q_j$ metrics, which have roughly the same tendency as $Q_h$ discussed above. We also show the neighborhood-hit values of the SDR results using data with varying SNR values ranging from 10 to 40 in the supplemental materials. All in all, Figures~\ref{fig:7qh_2d_synthetic} and~\ref{fig:3fdr_all_synthetic} show that our sharpening yields well-separated visual clusters but is less effective for data with sub-cluster structure. Improvements aimed at sub-cluster data are discussed in Section~\nameref{sec:discussion}.

\noindent\textbf{Real-world data:}
Figure~\ref{fig:5qm_banknote} shows $Q_h$, $Q_t$, $Q_c$, and $Q_j$ for different $k$-values measured on the real-world Banknote data set. Although $Q_t$, $Q_c$, and $Q_j$ yield higher values for LMDS than SLMDS, the projection results (Figure~\ref{fig:4fdr_all_realworld}) show that LMDS achieves a worse cluster separation compared with SLMDS. Even for \emph{t}-SNE, $Q_c$ and $Q_j$ yield higher values compared with S\emph{t}-SNE, but S\emph{t}-SNE exhibits a better cluster separation than \emph{t}-SNE (Figure~\ref{fig:4fdr_all_realworld}, second row).

Figure~\ref{fig:8qh_realworld} shows $Q_h$ measured for our four real-world data sets for different $k$-values. For the Olive oil data set, we use its super-class labels to compute $Q_h$. For this data set, it is important to limit $k$ since its classes are quite unbalanced: 323 (blue), 98 (orange), and 151 (yellow) points, respectively. Hence, we limit $k<300$ for this data set. Overall, Figure~\ref{fig:4fdr_all_realworld} shows that $Q_h$ decreases with $k$ for all studied methods. This is expected and in line with Figure~\ref{fig:4fdr_all_realworld}: As $k$ increases, $Q_h$ considers larger neighborhoods including points outside any visible (sub)cluster. The values of $Q_h$ for SLMDS are higher than for LMDS for all four data sets. These results reflect that the clusters are separated better in the sharpened projections than the original projections shown in Figure~\ref{fig:4fdr_all_realworld}. For \emph{t}-SNE and S\emph{t}-SNE, the results vary among different data sets. For the Banknote data, $Q_h$ for S\emph{t}-SNE is larger than for \emph{t}-SNE, which is also reflected in the projections shown in Figure~\ref{fig:4fdr_all_realworld}. However, for the other three data sets, \emph{t}-SNE shows higher $Q_h$-values than S\emph{t}-SNE. For the WiFi data, the results can be explained by the corresponding projections in Figure~\ref{fig:4fdr_all_realworld}, where \emph{t}-SNE mixes points from different clusters. For the Olive oil data, $Q_h$ considers neighbors outside a cluster with the same class labels for each divided cluster shown in Figure~\ref{fig:4fdr_all_realworld}(e)--(f). For the HAD data set, the $Q_h$-value for \emph{t}-SNE is slightly higher than that for S\emph{t}-SNE when $k<400$, but the situation reverses when $k \geq 400$. This can be explained by the significant oversegmentation exhibited in the projections (last row of Figure~\ref{fig:4fdr_all_realworld}(e)--(f)).

Figure 8 (supplementary material) complements Figure~\ref{fig:4fdr_all_realworld} and the discussion above by showing the $Q_c$, $Q_t$, $Q_j$, and $Q_{h}$ for the WiFi, Olive Oil, and HAD real-world data sets, and confirms that HD-SDR, while yielding better visual cluster separation than the DR baseline, scores slightly lower quality metrics.

Previous work using $Q_t$, $Q_c$, and $Q_j$ showed inconsistent results for different values of $k$, which resulted in interpretation difficulties\,\cite{trustworthiness_orig, jaccard_martins2015}. This is also visible in Figure~8 (supplemental materials): $Q_j$ increases with $k$, which is logical -- in the limit, when $k$ equals the sample count $N$, the neighborhood becomes the entire data set, so $Q_j=1$. $Q_t$ and $Q_c$ exhibit even more complex and non-monotonic behavior, often exhibiting local maxima for certain $k$-values.

In contrast to the above, $Q_h$ decreases monotonically with $k$: for small $k$-values, $Q_h$ has quite high values. This is expected for data sets that we know that are well separated into clusters having different labels (like ours). For such data sets, as long as $k$ is under the size of a cluster, $Q_h$ will be very high \emph{and} nearly constant, since a neighborhood will tend to `pick' same-label points from a single cluster. When $k$ exceeds the average number of samples having the same label (the average cluster size for data sets that are well separated into clusters), a neighborhood will inevitably contain more labels, resulting in $Q_h < 1$. In the limit, for a balanced data set of $C$ classes, $Q_h = 1/C$ when $k=N$. 
For all its limitations, $Q_h$ also has some advantages: $Q_h$ removes the dependency on the distance in the original data space. This is important as distances in that space are subject to the well-known dimensionality curse. As such, whenever Euclidean distances are used and the dimensionality increases, neighborhoods become meaningless or very unstable (the ratio of the closest and farthest points tends to one)\,\cite{Aggarwal01}. As $Q_h$ does not \emph{explicitly} check where neighborhoods in $n$D and 2D are the same, but only the homogeneity of \emph{labels} in a 2D neighborhood -- assuming again that these are homogeneous in the data space for a data set well-separated into clusters having different labels -- $Q_h$ is less sensitive to the above dimensionality issues.

Summarizing the above, we argue that although HD-SDR produces, in general, lower quality metrics than some of the baseline DR methods, (a) these quality metrics do not \emph{directly} capture the visual cluster separation we aim to optimize for; and (b) this separation actually is shown to \emph{increase} in the actual HD-SDR projections as compared to the baseline projections (see Figure~\ref{fig:4fdr_all_realworld}). However, visual cluster separation does not increase when oversegmentation occurs. The oversegmentation issues are discussed next in Section~\nameref{sec:discussion}. Based on the qualitative and quantitative studies above, we note that conducting user studies on SDR and comparing them with the quantitative results can be interesting for future work.

\section{Application to astronomical data}
\label{sec:astro}

\begin{figure*}[htb]
  \centering
  \includegraphics[width=0.9\linewidth]{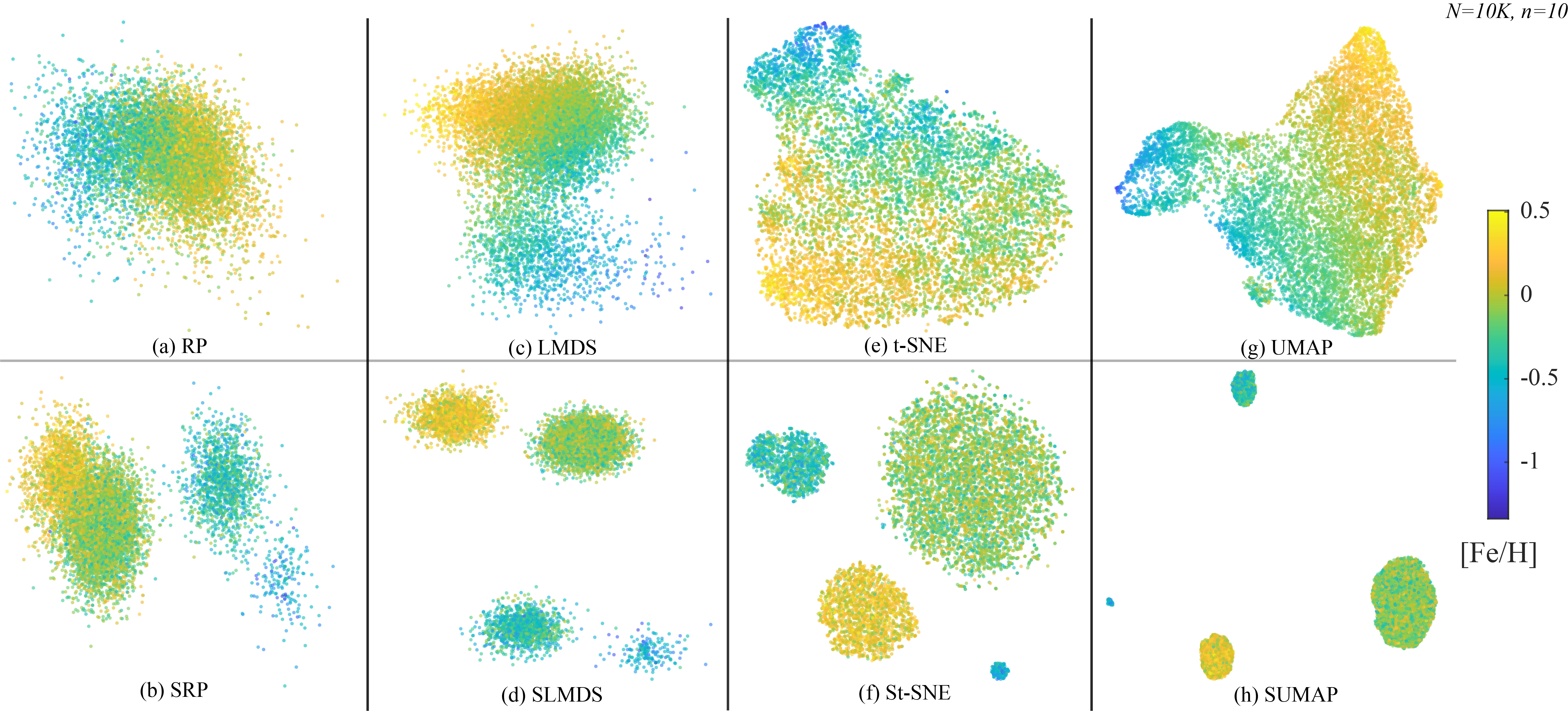}
  \parbox[t]{1\columnwidth}{\relax}
  \caption{\label{fig:9astro_fdr} We compare DR and HD-SDR (RP, LMDS, \emph{t}-SNE, and UMAP) for an unlabeled astronomical data set with no ground-truth labels: GALAH DR2 ($N=10\mathrm{K}$ with $n=10$). The projections are color-coded by one of the input values, [Fe/H], so that astronomers can further analyze the data\,\cite{tsne:astro}. The learning rate parameter is set to 0.18. Note that in all cases, HD-SDR shows a clearer separation of clusters compared with DR, and SLMDS, S\emph{t}-SNE, and SUMAP exhibit four major clusters with similar distributions of colors within each cluster, while SRP shows three major clusters with one of them having subclusters.}
\end{figure*}

As a specific use case to show that VCS improves by our HD-SDR method, we aim to separate 10K previously unclassified stars into clusters that may represent distinct physical groups within our own Milky Way galaxy. This is a common goal in astronomy: a large data set of unlabeled objects -- up to a few $10^9$ stars in current catalogues -- needs to be classified into separate (physically meaningful) clusters, so that labels representing physical groupings can be applied to individual objects. Importantly, this process has to involve the user in deciding which similar objects (in the same cluster) can be assigned to the same label. As such, the goal here is to perform \emph{manual} labeling, with labels having user-assigned semantics, and not automatic labeling of the type that clustering algorithms would support. Doing this \emph{manual labeling} object-by-object is clearly impossible with such large data sets. Previous attempts using standard DR methods have not been completely successful\,\cite{tsne:astro}. We show here that the VCS of HD-SDR meets this goal.

We first aim to reproduce the results shown in a recent study of dissecting stellar abundance space with \emph{t}-SNE\,\cite{tsne:astro} but using two more-recent data sets. First, we consider the second release of data from the Gaia satellite (known as Gaia DR2, publicly available since 2018) which contains observations of roughly 1.69 billion objects (stars, galaxies, quasars, and Solar System objects)\,\cite{astro:GAIADR2_1,astro:GAIADR2_2}. Secondly, we consider the second data release of the GALactic Archaeology with HERMES survey (GALAH DR2), also from 2018, a large-scale spectroscopic stellar survey including the properties of 342,682 stars in that release\,\cite{astro:GALAHDR2}. 

\par The data set we use cross-matches GALAH DR2 with Gaia DR2 using the Gaia DR2 ID of each star as the matching key. This cross-match yields 6D phase-space coordinates (3D stellar positions and 3D velocities). To obtain credible data, samples that meet the following criteria are excluded: one or more of positions $x$, $y$, and $z$ exceeding 25K parsec (where distance information becomes seriously unreliable), samples with missing values of any attribute (stellar abundance measurements and errors and 6D phase-space coordinates), and any stellar abundance measurements that are deemed by the GALAH team as unreliable\,\cite{astro:GALAHDR2}. From the 76270 credible samples, we randomly select $N=10\mathrm{K}$ samples to project using RP, LMDS, \emph{t}-SNE, and UMAP, and their sharpened versions. We use the same $n=10$ attributes, \emph{i.e.}, the stellar abundances [Fe/H], [Mg/Fe], [Al/Fe], [Si/Fe], [Ca/Fe], [Ti/Fe], [Cu/Fe], [Zn/Fe], [Y/Fe], and [Ba/Fe], 
as in a similar data set visualized by \emph{t}-SNE\,\cite{tsne:astro}, for comparison purposes.

\begin{figure}[htb]
  \centering
  \includegraphics[width=0.9\linewidth]{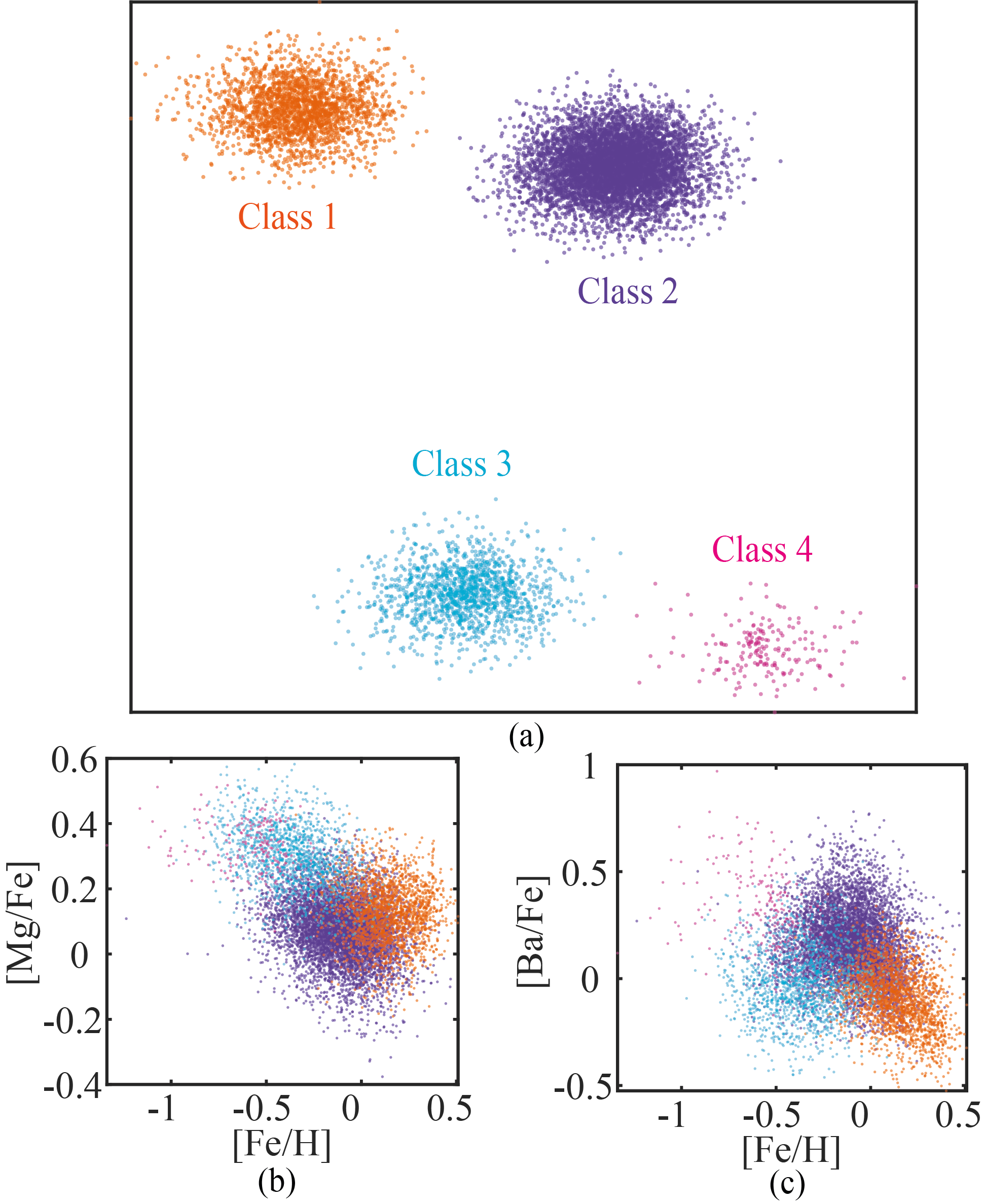}
  \parbox[t]{0.9\columnwidth}{\relax}
  \caption{\label{fig:10astro_tinsley} (a) SLMDS projection of the GALAH DR2 sample with clusters visually labeled by one of us. The labeled clusters in SLMDS help domains experts to further analyze the data as follows: (b) Tinsley diagram\,\cite{Tinsley80} shows the abundance of magnesium as a function of the iron abundance, used to interpret the origin and location of Milky Way stars. This diagram suggests to our domain-expert that stars in class 2 belong to the Milky Way's ``thin disk'', while those in classes 1 and 3 appear to belong to the Milky Way's ``metal-rich thick disk'' and ``metal-poor thick disk'', respectively; stars in class 4 appear to belong to the Milky Way's ``stellar halo''. (c) This plot shows the barium abundance of the stars as a function of their iron abundance, a tracer of a different nucleosynthetic process (the slow-neutron-capture, ``$s$-'', process). Stars in class 4 have strongly different barium abundances for their low iron content. These may be ``metal-poor barium stars'', which arise from binary star interactions. The same analysis for S\emph{t}-SNE and SUMAP are shown in the supplemental results; the clusters in SRP Figure~\ref{fig:9astro_fdr}(b) are not easily separated, thus excluded.}
\end{figure}

\par We present the resulting projections using HD-SDR and DR in Figure~\ref{fig:9astro_fdr}(a)--(h). The projections are color-coded by one of the input values, [Fe/H], so that astronomers can further analyze the data as in\,\cite{tsne:astro}; as a first pass, one of us examined the projections to evaluate the impact of the sharpening on understanding the astrophysical importance of the resulting distributions. Several insights follow. First, (without considering the color-coding) we can see that all HD-SDR results have better cluster separation compared with the DR results, and that SLMDS S\emph{t}-SNE, and SUMAP exhibit four major clusters with similar distributions of colors within each cluster. We note that our \emph{t}-SNE projection is very similar to that of Anders \emph{et al.}\ (compare Figure~\ref{fig:9astro_fdr}(e) with Figure 2 in\,\cite{tsne:astro}\footnote{The figures are similar but not identical because Anders \emph{et al.}\ used a different set of stellar abundances, the HARPS-GTO sample, based on stars taken for exoplanet identification, with ten times fewer stars but much higher quality; see\,\cite{tsne:astro} for more details.}). We also see that SRP exhibits three major clusters, with one of them having subclusters, as shown in panel (b). Overall, HD-SDR offers a much better cluster separation, even more so than \emph{t}-SNE, which is used to analyze a similar data set by Anders \emph{et al.}. We use one of the attributes, \emph{i.e.}, [Fe/H] to color-code the projection. By doing so, the clusters in HD-SDR projections are more easily explained by this attribute than the structures apparent in the DR projections.

\begin{figure}[htb]
  \centering
  \includegraphics[width=0.9\linewidth]{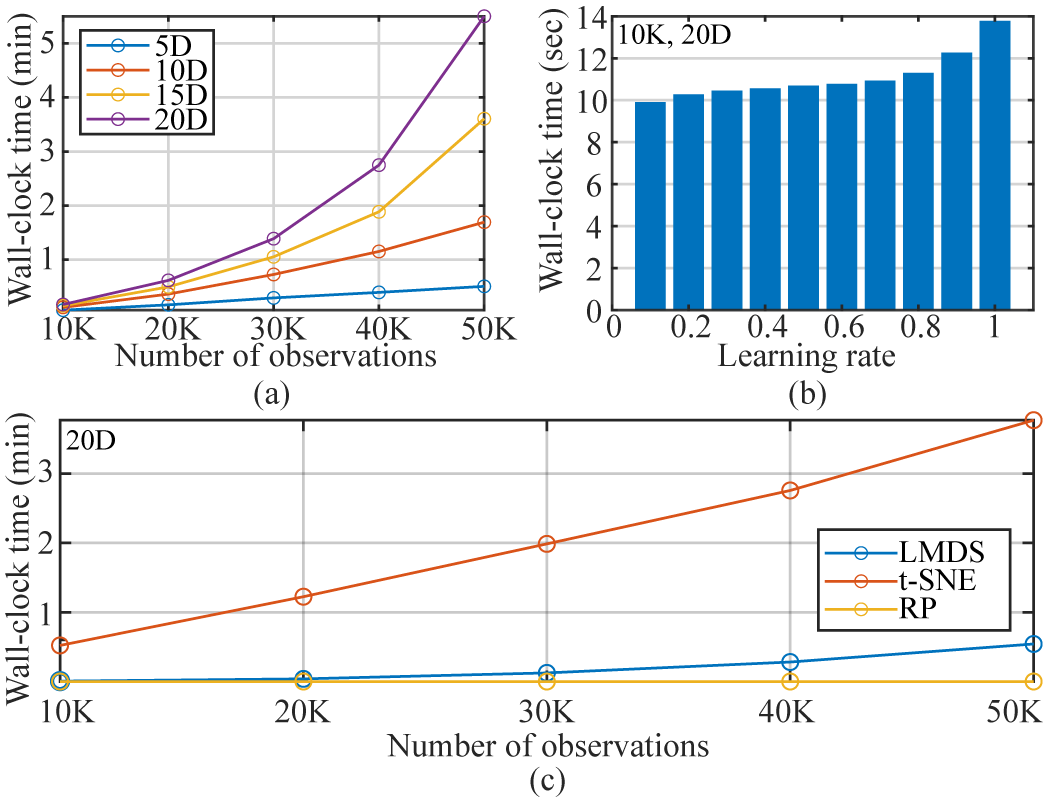}
  \parbox[t]{0.9\columnwidth}{\relax}
  \caption{\label{fig:11speed} (a) Wall-clock timing of LGC, for different number of observations along the $x$-axis and varying dimensions up to 20D. (b) Wall-clock time measurement of LGC on a 10$\mathrm{K}$ (20D) data set, using different learning rates $\alpha$. (c) Wall-clock time measurements of RP, LMDS, and \emph{t}-SNE applied to $n$D LGC data. We note that the wall-clock time of LGC increases with increasing $\alpha$ and depends heavily on the number of dimensions. Moreover, RP and LMDS (landmark ratio=0.05) take less than a minute to run, while \emph{t}-SNE takes longer and the speed heavily depends on the number of samples. More results of speed experiments using different numbers of dimensions in RP, LMDS, and \emph{t}-SNE and landmark ratios in LMDS are found in the supplemental materials.}
\end{figure}

Out of the four HD-SDR projections shown in Figure~\ref{fig:9astro_fdr}, we further analyze the projection using SLMDS of this data set and 2D scatter plots of three abundances (Tinsley diagram\,\cite{Tinsley80}) with [Fe/H], [Mg/Fe], and [Ba/Fe]) in Figure~\ref{fig:10astro_tinsley}. Note that the same analyses have been shown for S\emph{t}-SNE and SUMAP in the supplemental materials (SRP has been excluded because the subclusters are inseparable). Due to the clear separation of four clusters using SLMDS, a domain-expert is able to manually assign four different class labels to the clusters, as shown in Figure~\ref{fig:10astro_tinsley}(a). Next, we color-code the Tinsley diagram by the newly acquired labels (Figure~\ref{fig:10astro_tinsley}(b)--(c)). Without the labels, domain-experts would have to \emph{manually} visit each point to further analyze each star. Using the color-coded points, domain experts are able to quickly infer the location and origin of each group of stars in the Milky Way.

Upon seeing these results, one of us and two other domain experts in astronomy\footnote{Dr.\ Sarah Martell, Project Scientist of the GALAH Survey and a co-author of\,\cite{tsne:astro2}; and Dr.\ Sara Lucatello, an expert on tracing the formation of the Milky Way through the abundances of its stars.} noted that HD-SDR has a clear and higher potential in helping them to infer new results about the data at hand, compared with DR, in which clusters are less separable and are not strongly correlated with specific attributes. For example, in this case, the four classes could be identified in other tracers of the Milky Way's history, like its dynamical structure\,\cite{Helmi20}.

\section{Discussion}
\label{sec:discussion}
In this section, we discuss several aspects of HD-SDR.


\subsection{Scalability}
\label{sec:sub:scalability:speed}
\subsubsection{Speed}
\label{sec:scal_speed}
%

\begin{figure}[htb]
    \centering
    \includegraphics[width=0.9\linewidth]{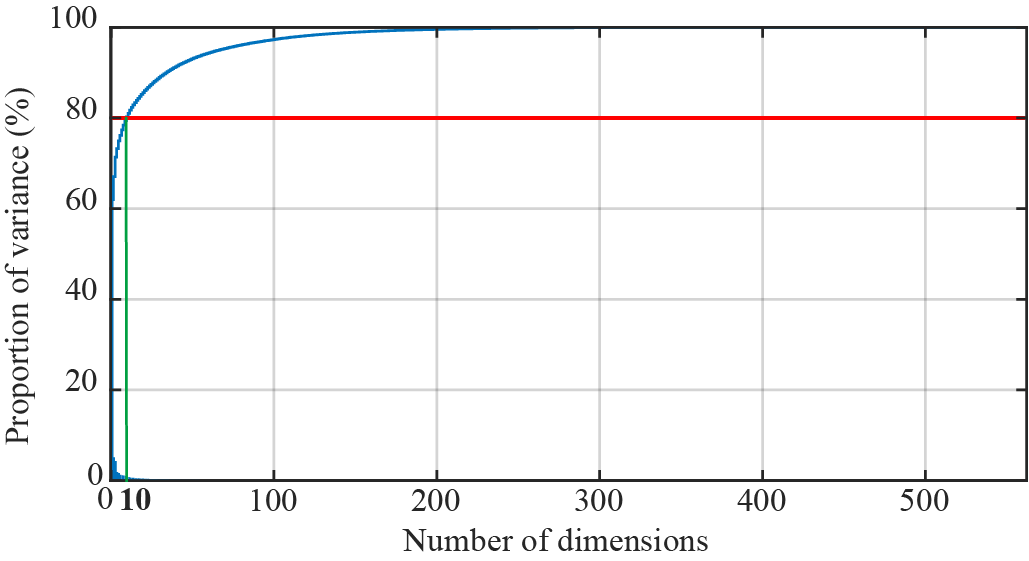}
    \parbox[t]{1\columnwidth}{\relax}
    \caption{\label{fig:pca_total_variance} Proportion of total variance explained by each component when using PCA on Human Activity Recognition (HAR) data. The HAR data set (561 dimensions) can be reduced to 10 dimensions while keeping 80\% of the variance.}
\end{figure}

\begin{figure}[htb]
    \centering
    \includegraphics[width=0.9\linewidth]{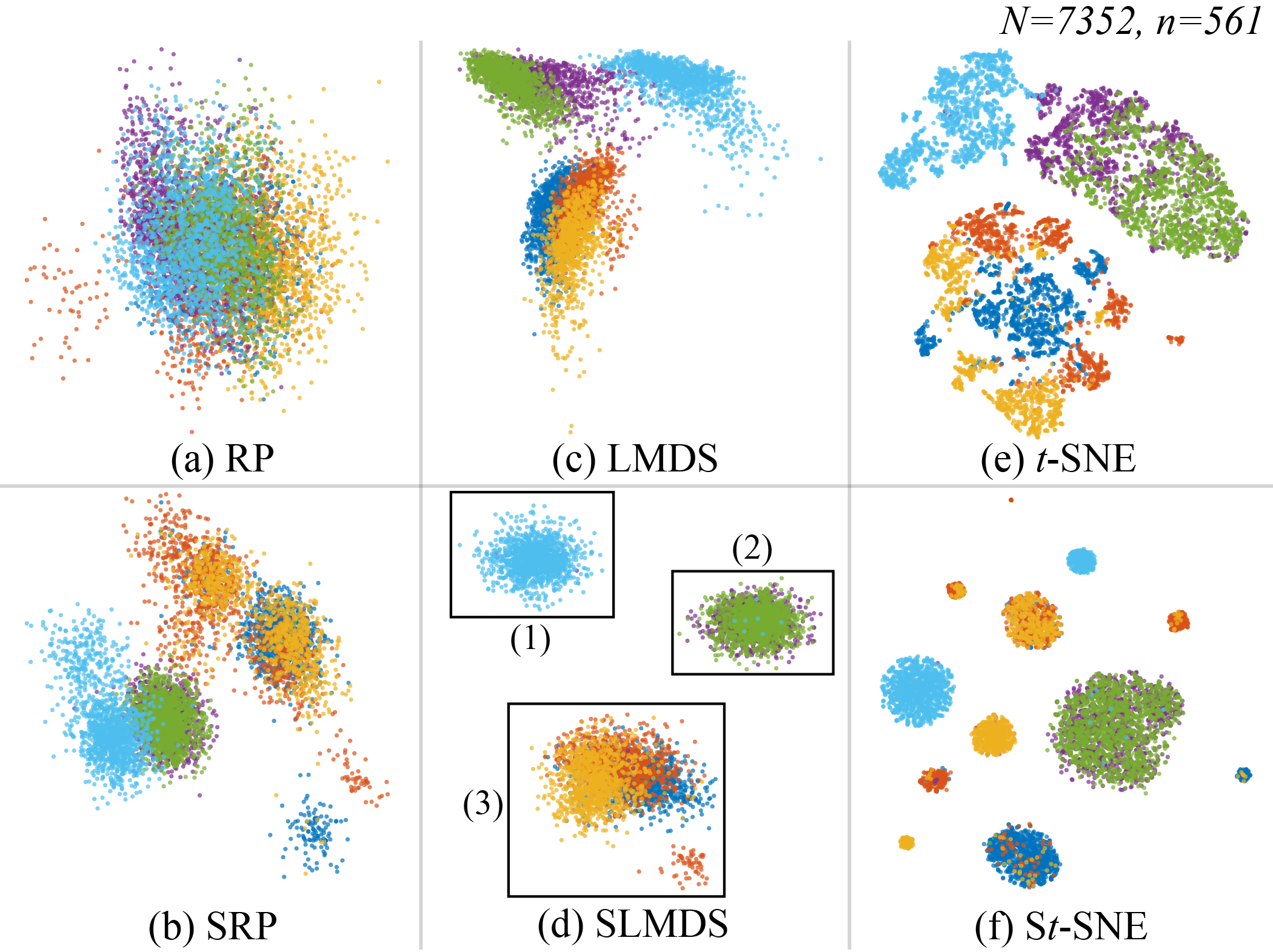}
    \parbox[t]{1\columnwidth}{\relax}
    \caption{\label{fig:pca_for_fdr_projections} Comparison of DR and HD-SDR on HAR data with dimensions pre-selected using PCA as in Figure~\ref{fig:pca_total_variance}. Note that the separation of clusters is slightly improved for HD-SDR as compared with DR (although subclusters are visually less separable). We note that nearly all samples in cluster (1) are from the \emph{lying} movement, cluster (2) mostly includes samples from static postures (standing and sitting), and cluster (3) mostly includes samples from the three dynamic activities.}
\end{figure}

Figure~\ref{fig:11speed}(a) shows the average wall-clock timings of LGC over 10 trials of randomly generated Gaussian data with five clusters for dimensionality $n \in \{5, 10, 15, 20\}$ and sample counts $N \in \{10\mathrm{K}, 20\mathrm{K}, 30\mathrm{K}, 40\mathrm{K}, 50\mathrm{K}\}$. For this plot, we used $\alpha=0.1$. LGC is mainly affected by $n$, due to the nearest neighbor search, which is of order $O(N\,n\,\log\,n)$. The overall time complexity of LGC is $O(n\,T\,N\,\log\,N)$. LGC takes over five minutes to compute for data with 20D and 50$\mathrm{K}$ samples (Figure~\ref{fig:11speed}(a)). Applying LGC to data with hundreds of dimensions may be impractical for end-users. A possible solution is outlined in Section~\nameref{sec:sub:scalability:speed:highD}.

Figure~\ref{fig:11speed}(b) shows how speed depends on the $\alpha$ parameter using a data set with $N=10\mathrm{K}$ samples and $n=20$. As $\alpha$ increases, the wall-clock time gradually increases. Profiling shows that this is due to the time needed to build $kd$-trees for kNN. About 95\% of the wall-clock time of LGC is due to the nearest neighbor (NN) search needed to compute $\rho$ (Eq.~\ref{Eq1}). Performing the kNN search for every iteration (instead of just once as in the original gradient clustering algorithm\,\cite{gc1975}) increases the time complexity proportional to the number of iterations ($T=10$) because $kd$-trees are constructed for every iteration. Easy speed-ups include replacing the current NN method\,\cite{nanoflann} by approximated and parallelized versions such as FLANN\,\cite{flann_2009,flann_2014}.
Multiple random projection trees (MRPT)\,\cite{mrpt:mrpt, mrpt:autoParamTuning} reduces expensive distance evaluations, thereby achieving higher speed than ANN and FLANN. However, MRPT has several issues: an insufficient number of requested nearest neighbors are returned; single-precision floating point is used; retrieving distances between neighbor points is not easily supported; and inaccurate search -- in the worst case, points that are far away from the query point are returned as nearest neighbors. Hence, MRPT is currently unfit for an accurate computation of nearest neighbors. Separately, the sample shift (Eq.~\ref{Eq2}) can be trivially parallelized on the CPU or GPU for further acceleration, leading to speed-ups of two orders of magnitude, as shown by related work\,\cite{cubu}.

Figure~\ref{fig:11speed}(c) shows the wall-clock timings for LMDS, \emph{t}-SNE, and RP on $n$D LGC data. When comparing the time measurements of LGC against standalone DR methods, all DR methods take less time to run compared with LGC for 50$\mathrm{K}$ observations, where \emph{t}-SNE takes the longest. Note that LMDS, \emph{t}-SNE, and RP are all from the same Tapkee library (UMAP is not and therefore has been excluded from the experiments). More timing results using different numbers of dimensions in RP, LMDS, and \emph{t}-SNE, and landmark ratios in LMDS, can be found in the supplemental materials.

\subsubsection{High-dimensional data}
\label{sec:sub:scalability:speed:highD}
Section~\nameref{sec:scal_speed} states that applying LGC to data with hundreds of dimensions may be impractical due to speed issues. A solution is to first reduce the dimensionality with a simple and fast DR (\emph{i.e.}, PCA) and then apply HD-SDR. Figures~\ref{fig:pca_total_variance}--\ref{fig:pca_for_fdr_projections} illustrate this. Here, Human Activity Recognition (HAR) data\,\cite{realworld:uciML,realworld:HAR:UCI} with $n=561$ and $N=7352$ for six basic activities are used: three static postures (standing, sitting, and lying); and three dynamic activities (walking, walking downstairs, and walking upstairs). First, we reduce $n$ to 10 using PCA (keeping 80\% of total variance, see Figure~\ref{fig:pca_total_variance}). Then, we use HD-SDR on this 10-dimensional data, with $\alpha=0.2$. The obtained projections using SRP, SLMDS, and S\emph{t}-SNE (Figure~\ref{fig:pca_for_fdr_projections} bottom row) all exhibit improvements over their original counterparts (Figure~\ref{fig:pca_for_fdr_projections} top row). In particular, we can easily see that cluster (1) of SLMDS (Figure~\ref{fig:pca_for_fdr_projections}(d)) is clearly separated from the others. Nearly all samples in this cluster are from the \emph{lying} movement class. Although points with different class labels are mixed in clusters (2) and (3), most points in cluster (2) are from other static postures (standing and sitting), while most points in cluster (3) are from the three dynamic activities. Further note that separating sub-clusters still remains an issue (see Section~\nameref{limitations} next).


\begin{figure*}[htb]
    \centering
    \includegraphics[width=0.95\linewidth]{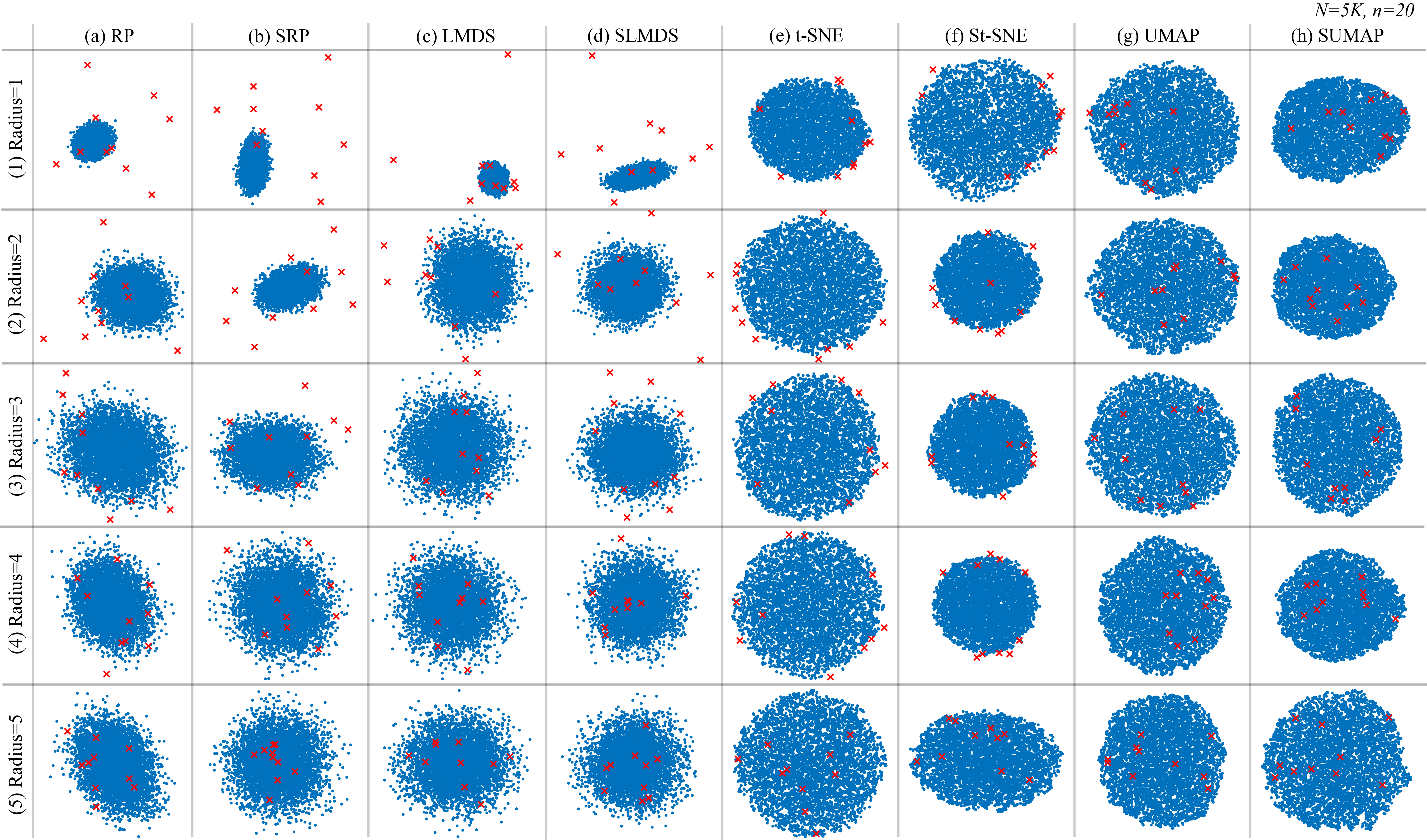}
    \parbox[t]{0.95\columnwidth}{\relax}
    \caption{\label{fig:added_2_outliers}Comparison of DR and HD-SDR using 10 outliers (red crosses) and 5$\mathrm{K}$ random points uniformly-distributed in a 20-dimensional hypersphere centered at the origin and with different radii $R \in \{1,2,3,4,5\}$. The distance between an outlier and the origin is always five. Note that \emph{t}-SNE and UMAP and their sharpened versions fail to preserve the outliers even when the distance between the outliers and the cluster is the largest (row 1). Further note that SRP and SLMDS preserve similar or larger numbers of outliers compared with RP and LMDS, respectively.}
\end{figure*}

\subsection{Data distortion}
Our method addresses the cluster separation problem by shifting points in the original space, which may cause data distortions. Section~\nameref{sec:evaluation_LGC} shows that the LGC step actually \emph{improves} neighborhood preservation with respect to the ground-truth labels in the original data, which is the main aspect we aim to capture. \emph{Any} DR method performs, by definition, non-trivial amounts of data distortion when mapping from the high-dimensional space to 2D if the data is not originally already located on a smooth 2D manifold. Hence, no DR method can faithfully capture \emph{all} aspects of any data set\,\cite{mateusDR_survey2019}. Users will be always exposed to certain types of data distortions and/or data aspects that are not captured in the 2D projection. This is especially true for local and nonlinear projection techniques, \emph{e.g.}, t-SNE and UMAP. Whether such distortions occur in the \emph{preprocessing} step like our LGC, or in the \emph{projection}, as for all other DR methods, does not remove the fact that such unavoidable changes occur. Hence, the fact that LGC changes the data does not imply that our technique is less trustworthy than \emph{any} other DR technique, which change the data during the projection itself.

\subsection{Relation to clustering}
Our technique is aimed at supporting the exploratory analysis using DR methods, and thus has a close relation to data clustering. For example, Chen \emph{et al.}\,\cite{realworld:banknote_sep} use mean shift, which is closely related to LGC, to create DR projections. They construct an \emph{explicit} clustering of a data set $D$, $\cup_i C_i = D$, by mean shift, after which they project the cluster centers $c(C_i)$ to 2D and use these landmarks $P(c(C_i))$ to perform \emph{local} MDS projections $P(C_i)$. Many other local DR methods work similarly\,\cite{lamp:original,nonato18}. In contrast, we do not require an explicit clustering of the data to partition $D$ to project it piece-wise; rather, we use LGC as a \emph{preconditioning} technique to improve a subsequent \emph{global} projection of $D$. Moreover, our LGC updates the KDE gradient $\nabla \rho$ at every advection iteration (Eq.~\ref{Eq2}). This is different from classical mean shift\,\cite{gc1975} as used in\,\cite{realworld:banknote_sep}, where $\nabla \rho$ is computed from the initial density estimate and then used unchanged during the update (Eq.~\ref{Eq2}). Updating $\nabla \rho$ leads to faster cluster separation, especially for noisy data\,\cite{alex1,cubu}.

Separately, as mentioned already in Section~\nameref{sec:relatedwork:VCS}, HD-SDR cannot, and should not, create projections with high VCS for \emph{all} datasets. This would be misleading as it would suggest to the user that such structures exist in otherwise unstructured data. Hence, for datasets that lack such structure, one should expect HD-SDR to create projections with low VCS.

\subsection{Preservation of outliers}
\label{sec:outliers}
Figure~\ref{fig:added_2_outliers} compares DR and HD-SDR using data sets with outliers marked as crosses. To investigate the effect of SDR on outliers, we create five 20-dimensional hyperspheres, with 5$\mathrm{K}$ points randomly generated and uniformly distributed within radii $R \in \{1,2,3,4,5\}$. We then add 10 outliers distributed on the largest hypersphere surface ($R=5$) to each data set. Unlike RP and LMDS, \emph{t}-SNE and UMAP and their sharpened versions do not preserve outliers well. This is expected because \emph{t}-SNE and UMAP are \emph{neighborhood}-preservation DR methods, whereas RP and LMDS are \emph{distance}-preservation methods (see Section~\nameref{sec:relatedwork:DRforLabeling}). As Figure~\ref{fig:added_2_outliers} (rows 1 and 2) show, the farther the outliers are from the hypersphere surface, the more outliers are preserved by SRP and SLMDS than by RP and LMDS. As the hypersphere radius increases, both DR and SDR do not preserve outliers well, which is expected because the distance between the outliers and the data in the sphere decreases (rows 4 and 5). Overall, SDR preserves a similar number of outliers compared with DR, but a careful selection of parameters is needed for real-world data sets where the number of clusters and outliers vary. Section~\nameref{sec:discussion:parameters} further discusses parameter setting.

\subsection{Limitations and future work}
\label{limitations}


\subsubsection{Undersegmentation and oversegmentation}
In some fields, including subfields of astronomy, domain experts are interested in finding \emph{substructures}\,\cite{astro:subclusters}, which relates to the \emph{undersegmentation} issue. As Figure~\ref{fig:3fdr_all_synthetic} shows, our method is less effective in capturing tightly connected subclusters. This is closely related to how far apart two clusters should be to be separated by the projection rather than rendered as a single cluster. This issue should be further explored both theoretically and empirically.

\begin{figure}[htb]
    \centering
    \includegraphics[width=0.9\linewidth]{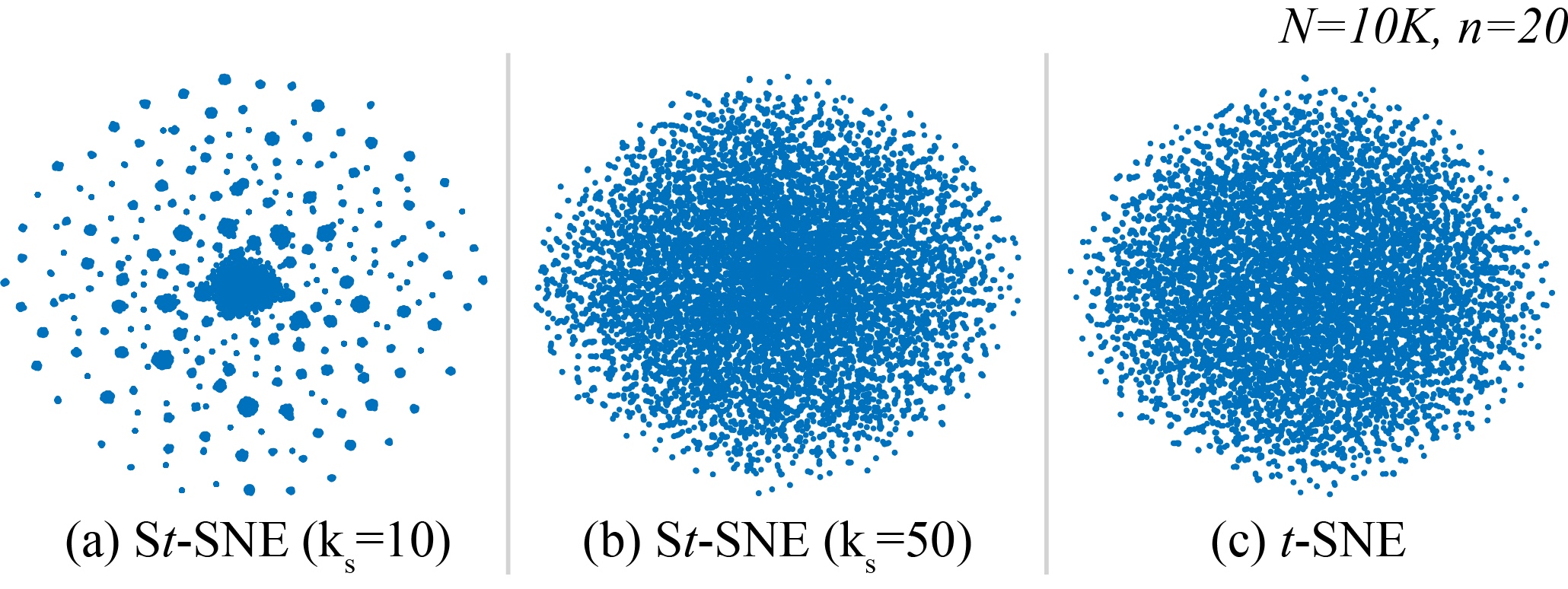}
    \parbox[t]{1\columnwidth}{\relax}
    \caption{\label{fig:added_1_randomNoise} Results of DR and SDR with varying $k_s$-values for sharpening. The synthetic data set consists of 10$\mathrm{K}$ randomly generated samples in 20D. Note that S\emph{t}-SNE shows oversegmented clusters using $k_s=10$, while (b) shows a single cluster as in (c) when using $k_s=50$.}
\end{figure}

While the subcluster issue can also be seen as undersegmentation, HD-SDR further augments oversegmentation when using an oversegmentation-prone DR or a DR with strong cluster separation like UMAP or \emph{t}-SNE (see Figure~\ref{fig:4fdr_all_realworld}, fourth row). Oversegmentation is a known aspect of mean shift methods\,\cite{alex1}. While there are many data-specific heuristics to set the scale of clusters, there is no generic way to avoid under- or oversegmentation. Hence, oversegmentation is not particular to our method, as other projections with strong clustering also exhibit this problem.

Besides controlling the parameters of our method, one way of solving under- and oversegmentation would be to use an adaptive learning rate $\alpha$ to capture different detail levels in a cluster. Adaptive learning rates that consider the distribution of distances between neighboring points may allow sharpening to be more adaptive to different distributions of clusters. However, this approach is also risky because of errors in the density estimator propagating due to inconsistent point-shifts after each iteration. Studying adaptive learning rates is hence left to future work.

\subsubsection{Noise-free data}
\label{sec:discussion:sharpenNoise}
When using SDR, it is assumed that there are no errors in the measurements or data processing stage. In real-world data sets, the data may intrinsically have uncertainties such as measurement errors or errors due to data processing. For example, with astronomical data, there can be measurement errors or missing values. Even though we show that our method is noise-resistant up to $\mbox{SNR}=10$ (synthetic data type (5)), real-world data may contain non-Gaussian noise, which is why it is crucial for the user to consider these uncertainties for a more accurate analysis of the data at hand.

Furthermore, random noise can be sharpened producing oversegmentation when using our method. However, it is possible to negate the effects of sharpening small dense areas of noise by using a large enough value of $k_s$ in SDR. We demonstrate an extreme case in Figure~\ref{fig:added_1_randomNoise} by comparing two different $k_s$-values in SDR using a single cluster with 10$\mathrm{K}$ randomly generated samples in 20D. We see that \emph{t}-SNE and S\emph{t}-SNE using $k_s=50$ show a single cluster, whereas S\emph{t}-SNE using $k_s=10$ shows highly oversegmented clusters. The same observation can also be made using RP, LMDS, and UMAP (see supplemental materials). Hence, it is important for the user to select a large enough $k_s$-value (i.e., $k_s \geq 50$) to prevent sharpening noise that may cause oversegmentation.

\subsubsection{Parameter setting}
\label{sec:discussion:parameters}
Our parameters $k_s$ (how localized a shift is) and $\alpha$ (shift speed) are interconnected, see Section~\nameref{sec:method:lgc}. This also holds for other kernel density estimation (KDE) methods\,\cite{alex1}. While both $k_s$ and $\alpha$ affect the segmentation degree, if $k_s$ is large enough, then larger $k_s$-values may not significantly affect segmentation without choosing an suitable $\alpha$-value, see Figure~\ref{fig:1params_gaussian} (second and third rows). Setting a large enough $k_s$-value is also crucial to avoid sharpening noise as explained in Section~\nameref{sec:discussion:sharpenNoise}, which is why we use $k_s \geq 50$.

Figure 9 (supplemental material) completes the insights from Figures~\ref{fig:1params_gaussian}--\ref{fig:2params_non_Gaussian} by showing the results of our method for the WiFi data set for multiple values of $\alpha$ and $k_s$, using $T=10$ iterations, as discussed in Section~\nameref{sec:method:lgc}. As explained in Section~\nameref{sec:data_traits} and also visible in Figure~\ref{fig:4fdr_all_realworld}, we know that this data set consists of four clusters. We see that HD-SDR produces four compact and well-separated clusters for the parameter combination $\alpha=0.15$ and $k_s=100$, which are in line with our recommended presets ($\alpha=0.15$, $k_s \geq 50$). The over- and undersegmentation produced by other parameter values follow the same trend for this real-world data set as for the synthetic data in Figure~\ref{fig:1params_gaussian}.

A too-large number of LGC iterations ($T$-value) can lead to over-shooting the local cluster centers during the gradient ascent and also longer computation. While Hurter \emph{et al.}\,\cite{alex1} decrease the advection speed $\alpha$ over iterations to solve the overshooting problem, they aim to have all points in a visual cluster converge to a \emph{single} location. This is clearly undesired for projections, so we use a constant advection speed. Finally, note that we stop LGC based on a fixed $T$-value. A better stop criterion would be to use a quality metric, \emph{e.g.}, neighborhood-hit ($Q_h$). Exploring this (and how to do it efficiently) would be interesting for future work.

\subsubsection{Post-processing in dimensionality reduction}
Technically, our sharpening approach can also be applied to 2D instead of $n$D data. However, this is problematic: We know in advance that the data distances are `uniform' in $n$D, so sharpening with a certain distance or speed will work uniformly for all data points\,\cite{gc1975, ms}. In contrast, in a 2D projection, distances are generally non-uniform -- one 2D pixel may correspond to small or large data distances depending on where it is in the projection. Hence, we cannot sharpen 2D points with the same speed, and determining the speed to use per point is a major difficulty, as this would require knowing the inverse projection $P^{-1}$. 
Another problem with sharpening after DR is that a poor projection (in terms of VCS) cannot possibly be `fixed' by sharpening; sharpening will make it worse, as it will amplify its poor VCS. In contrast, sharpening before DR can produce clear VCS even for DR methods that originally exhibit poor VCS as shown in Figures~\ref{fig:4fdr_all_realworld} and \ref{fig:9astro_fdr}.

Other post-processing methods for 2D projections exist, \emph{e.g.}, applying `clustering' after DR or SDR to automatically label individual clusters in a projection. This approach is currently out of our scope and will be explored in future works.

\subsubsection{LGC for \emph{t}-SNE and UMAP}
Figures~\ref{fig:3fdr_all_synthetic}--\ref{fig:4fdr_all_realworld} show that some DR methods produce a clearer cluster separation even without LGC, in particular methods that already exhibit strong cluster separation and/or show oversegmentation, \emph{e.g.}, \emph{t}-SNE and UMAP. UMAP yields a better VCS than other SDR results including SUMAP in most of the examples shown in this paper except for Figure~\ref{fig:9astro_fdr}. However, due to the limitations of UMAP mentioned in Section~\nameref{sec:relatedWork} (too dense clusters and difficult parameter setting due to its stochastic nature), other DR methods with lower VCS may be preferred over UMAP, which is why we explore the sharpening effect on additional DR methods. This is also why we explicitly compare SDR with the original DR methods rather than with specific DR methods like UMAP or t-SNE.

\subsubsection{Selection of baseline DR method}
Figures~\ref{fig:3fdr_all_synthetic}--\ref{fig:4fdr_all_realworld} show that some DR methods yield a clearer cluster separation when aided by LGC than other methods. Besides these examples, other DR methods benefit from being combined with LGC. To study this, we applied LGC to all 44 DR methods in the benchmark of Espadoto \emph{et al.}\,\cite{mateusDR_survey2019} for the WiFi data set using $\alpha=0.15$ (see supplemental materials). These results show that our method works with any DR method that we are aware of. While we use the same $\alpha=0.15$ for all experiments, some combinations of LGC with certain DR methods produce better results with different $\alpha$-values. In particular, ISO and LMVU produce some separation of clusters, while the sharpened versions of them do not. Using a smaller $\alpha$ for S-ISO and S-LMVU results in clearer cluster separation (see supplemental materials). This suggests that using different $\alpha$-values can solve issues with poorer cluster separation in HD-SDR than in DR. While out of scope of this paper, exploring why certain DR methods are more suitable for HD-SDR based on a user-centric approach\,\cite{perceptionChooseDR_userstudy2}, and which SDR is effective for different data types, are important topics to study next.

\section{Conclusion}
\label{sec:conclusion}
We have presented a new method for dimensionality reduction (DR) that creates visually separated sample clusters targeted for user-guided labeling to explore and analyze the data using DR. Key to our method is a preconditioning step that ``sharpens'' the sample density in the data space prior to using \emph{any} DR technique. We tested our method using both synthetic and real-world data from five different application domains, using RP, LMDS, \emph{t}-SNE, and UMAP as DR methods. HD-SDR yields better visual cluster separation in the projection than the original DR methods that exhibit weak cluster separation. In terms of practical usefulness, astronomy experts see clear added-values in the results produced by HD-SDR on their data as compared with \emph{t}-SNE, which was used in previous studies. This is the first time to our knowledge that a mean shift-based sharpening method is used without any prior knowledge of cluster modes to enhance the separability of clusters. We suggest that the LGC preconditioning step shows how LGC can lead to techniques that bridge the gap between DR methods with different abilities to separate clusters.

\begin{acks}
This work is supported by the DSSC Doctoral Training Programme co-funded by the Marie Sklodowska-Curie COFUND project (DSSC 754315). The GALAH survey is based on observations made at the Australian Astronomical Observatory, under programmes A/2013B/13, A/2014A/25, A/2015A/19, A/2017A/18. We acknowledge the traditional owners of the land on which the AAT stands, the Gamilaraay people, and pay our respects to elders past and present. This work has made use of data from the European Space Agency (ESA) mission \emph{Gaia} (\url{https://www.cosmos.esa.int/gaia}), processed by the \emph{Gaia} Data Processing and Analysis Consortium (DPAC, \url{https://www.cosmos.esa.int/web/gaia/dpac/consortium}). Funding for the DPAC has been provided by national institutions, in particular the institutions participating in the \emph{Gaia} Multilateral Agreement.
\end{acks}

\bibliographystyle{SageV}
\bibliography{template.bib}

\end{document}